\documentclass[conference]{IEEEtran}

\usepackage[english]{babel}
\usepackage[T1]{fontenc}
\usepackage[utf8]{inputenc}
\usepackage{amsmath,amssymb,amsfonts}
\usepackage{mathtools}          % après amsmath
\usepackage{booktabs}
\usepackage{array}
\usepackage{tabularx}
\usepackage{multirow}
\usepackage{graphicx}
\usepackage{float}
\usepackage[font=footnotesize,labelsep=period,justification=justified,singlelinecheck=true]{caption}
\usepackage{subcaption}
\usepackage{cite}
\usepackage{longtable}          % inutilisable en deux colonnes (non utilisé ici)
\usepackage{etoolbox}           % \AtBeginEnvironment
\usepackage{fancyvrb}
\usepackage{upquote}
\usepackage{enumitem}
\usepackage{newunicodechar}

\usepackage{newtxtext,newtxmath}
\usepackage{siunitx}
\usepackage{microtype}          % APRÈS les paquets de polices

\usepackage{xcolor}
\definecolor{accentblue}{HTML}{2E5C8A}
\definecolor{accentgreen}{HTML}{2E8B57}
\definecolor{accentred}{HTML}{C0392B}
\definecolor{boxfillneutral}{HTML}{F2F2F2}
\definecolor{boxfillgreen}{HTML}{E6F4EA}
\definecolor{boxfillred}{HTML}{FBEAE9}
\definecolor{boxfillblue}{HTML}{E8F0F9}

\usepackage{tikz}
\usetikzlibrary{arrows.meta,positioning,calc,fit,shapes.geometric,backgrounds,decorations.markings}
\usepackage{pgfplots}
\pgfplotsset{compat=1.18}
\usepgfplotslibrary{colorbrewer}
\pgfplotsset{every axis/.append style={line width=0.6pt}}

\usepackage[colorlinks=true, linkcolor=black, citecolor=black,
            urlcolor=accentblue]{hyperref}

\newunicodechar{−}{\ensuremath{-}}
\newunicodechar{≥}{\ensuremath{\geq}}
\newunicodechar{≤}{\ensuremath{\leq}}
\newunicodechar{√}{\ensuremath{\surd}}
\newunicodechar{₂}{\ensuremath{{}_2}}
\DeclareUnicodeCharacter{00B2}{\ensuremath{{}^2}}   % \texttwosuperior absent de ntxtlf (R² ...)

\newcolumntype{Y}{>{\centering\arraybackslash}X}
\newcolumntype{L}{>{\raggedright\arraybackslash}X}
\AtBeginEnvironment{longtable}{\footnotesize}
\begin{document}

\title{Data Quality Audit and Bivariate Temperature--Humidity Analysis of a Long-Term LoRaWAN Network: Anomaly Detection, PCA, Clustering and Machine Learning}

\author{
    \IEEEauthorblockN{Mamadou SOW}
    \IEEEauthorblockA{LIPN-CNRS7030, Universit\'e Sorbonne Paris Nord\\
    99 avenue JB Cl\'ement, F-93430 Villetaneuse, France\\
    mamadou.sow@univ-paris13.fr}
}

\maketitle

\begin{abstract}
Long-term environmental IoT/LPWAN deployments often aggregate heterogeneous data streams from diverse devices without systematic auditing of each variable's actual coverage prior to analysis---an increased risk when multiple sensor types coexist within the same network. This work proposes a C++17 processing pipeline applied to the Tour Perret LoRaWAN dataset (421{,}937 frames, June 2021--June 2023, 5 devices). A systematic coverage audit integrated into the extraction phase reveals that CO$_2$ is completely missing (0\% coverage) and pressure is reported only by an isolated base module (Wyres, 4.24\% of the corpus), invalidating the initial hypothesis of a joint four-variable analysis. The analysis is refocused on the Temperature--Humidity pair, jointly available across 205{,}005 frames (95.76\%) from 4 Elsys EMS sensors. The pipeline sequentially performs anomaly detection (modified Z-score, Isolation Forest), bivariate Principal Component Analysis, clustering (K-means, DBSCAN), and a multilayer perceptron predicting humidity from temperature and time, whose residuals serve as a detector for slow drift. On real-world data: Temperature--Humidity correlation $r=-0.662$, two-component PCA (PC1 $=83.1\%$ of variance), Isolation Forest detects $2.12\%$ multivariate anomalies while univariate Z-score detects none, calibrated DBSCAN isolates $0.17\%$ noise across 13 clusters, and the MLP achieves $R^2=0.672$ (RMSE $=10.54\%$) in cross-validation. This work demonstrates that per-variable coverage auditing prior to modeling is a necessary methodological step for heterogeneous multi-sensor IoT datasets.
\end{abstract}


\section{Introduction \& Coverage Audit}\label{sec:intro}

\subsection{Context and Operational Challenges}\label{sub:context}
Long-term environmental monitoring via Low-Power Wide-Area Networks
(LPWAN)---specifically using the LoRaWAN protocol---presents unique data
engineering and data quality challenges. In real-world deployments such
as the Perret Tower (\emph{Tour Perret}) dataset \cite{donsez2023tour},
raw telemetry streams suffer from network packet loss, heterogeneous
payload formats, multi-gateway frame duplication, and hardware
configuration shifts over time.

A common pitfall in multivariate anomaly detection and predictive
modeling is the \emph{a priori} assumption of variable co-occurrence.
Researchers and practitioners often design processing pipelines
expecting all target variables---such as Temperature (\(T\)), Relative
Humidity (\(H\)), Atmospheric Pressure (\(P\)), and Carbon Dioxide
concentration (\(\mathrm{CO_2}\))---to be present synchronously in every
transmitted frame. However, in multi-sensor LPWAN deployments, distinct
physical sensors often operate on different transmission schedules,
report varying payload schemas across firmware revisions, or belong to
completely separate device applications aggregated under a single
network server instance.

% Spans two columns
\begin{figure*}[t]
\centering
\resizebox{0.9\textwidth}{!}{%
\begin{tikzpicture}[
  node distance=10mm and 10mm,
  box/.style={draw, rounded corners=2pt, minimum width=5.3cm, minimum height=1.15cm,
              align=center, font=\small, line width=0.8pt, fill=boxfillneutral},
  boxgreen/.style={box, fill=boxfillgreen, draw=accentgreen},
  boxred/.style={box, fill=boxfillred, draw=accentred},
  boxblue/.style={box, fill=boxfillblue, draw=accentblue},
  arr/.style={-{Latex[length=2.2mm]}, line width=0.9pt, draw=gray!60!black}
]
\node[box] (raw) {\textbf{Raw LoRaWAN Telemetry Stream}\\\footnotesize (Aggregated JSON-Lines Log Files)};
\node[boxgreen, below=of raw] (audit) {\textbf{1. Coverage Audit}\\\footnotesize Evaluate schema completeness by Variable \& devEUI};
\node[boxred, below left=14mm and 4mm of audit] (excl) {\textbf{Excluded Variables}\\\footnotesize CO$_2$: Totally absent (0\%)\\\footnotesize Pressure: Structurally isolated\\\footnotesize (Wyres stream)};
\node[boxblue, below right=14mm and 4mm of audit] (ret) {\textbf{Retained Bivariate Core Dataset}\\\footnotesize Elsys EMS frames (95.76\%)\\\footnotesize Co-occurring T \& H\\\footnotesize $n=205{,}005$ complete frames};
\coordinate (midlow) at ($(excl.south)!0.5!(ret.south)$);
\node[box, below=14mm of midlow, anchor=north] (clean) {\textbf{2. Preprocessing \& Cleaning}\\\footnotesize Deduplication (devEui $\vert$ fCount)\\\footnotesize Physical bounds filtering};
\draw[arr] (raw) -- (audit);
\draw[arr] (audit) -- (excl);
\draw[arr] (audit) -- (ret);
\draw[arr] (ret.south) |- ($(ret.south)+(0,-7mm)$) -| (clean.north);
\end{tikzpicture}%
}
\caption{LoRaWAN telemetry ingestion and cleaning pipeline: the coverage audit (Step 1) is executed before any downstream analysis, separating structurally unusable variables (CO$_2$, pressure) from the retained bivariate (T,H) core dataset.}
\label{fig:lorawan_pipeline_audit}
\end{figure*}
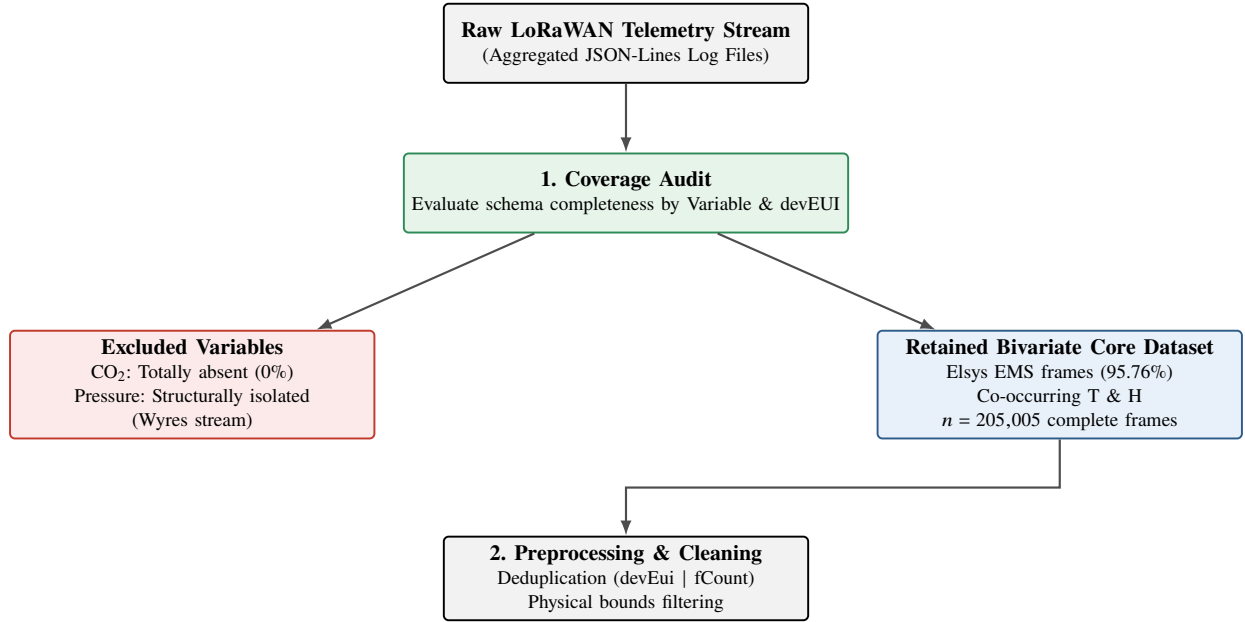

\subsection{Central Finding: Systematic Data Coverage Audit}\label{sub:central}

To prevent silent pipeline failures and catastrophic data distortion
(such as invalid outer joins producing null-dense matrices or artificial
cross-device frame alignment), an explicit, automated \textbf{Data
Coverage Audit} was embedded directly into the primary C++ extraction
module (\texttt{01\_extract.cpp}). The audit evaluated payload
completeness across all registered device identifiers (\texttt{devEUI})
and application boundaries (\texttt{applicationName}).

The coverage audit revealed two critical structural properties of the
raw dataset:
\begin{enumerate}

    \item \textbf{Total Absence of Carbon Dioxide (\(\mathrm{CO_2}\)):} Despite
  initial project documentation indicating \(\mathrm{CO_2}\) monitoring
  for indoor environmental quality, zero frames across the multi-year
  corpus contained decoded \(\mathrm{CO_2}\) payload fields. This stems
  either from a physical deployment omitting the \(\mathrm{CO_2}\) gas
  sensor module or an unupdated network decoder instance.
  \item \textbf{Structural Isolation of Barometric Pressure (\(P\)):} Pressure
  measurements originated exclusively from a secondary device cohort
  (Wyres sensors), whereas the primary sensor fleet consisted of four
  Elsys EMS (Environmental Monitoring Sensor) nodes. Because the Wyres
  and Elsys EMS devices transmitted asynchronously without clock
  synchronization, barometric pressure was structurally disconnected
  from the main stream at the individual frame level.

\end{enumerate}

\subsection{The Bivariate Temperature--Humidity Dataset}\label{sub:bivariate}

Attempting an inner join across all four initial variables
(\(T, H, P, \mathrm{CO_2}\)) would yield \textbf{0 usable rows}, causing
a complete execution failure. Conversely, performing a naive
cross-device alignment would artificially merge environmental readings
from physically distinct locations within the tower structure.

Consequently, the scope of the study was refocused on the largest,
physically consistent, co-occurring payload schema: the
\textbf{bivariate Temperature--Humidity (\(T, H\)) feature space}
provided by the four Elsys EMS nodes.

{\sloppy
Following extraction, ISO 8601 timestamp parsing to Unix epoch,
multi-gateway frame deduplication (keyed on
\texttt{devEui\textbar{}fCount}), and physical boundary validation
(\(T \in [-40, 85]\text{ °C}\), \(H \in [0, 100]\%\)), the clean corpus
comprised \textbf{\(n = 205{,}005\) complete bivariate frames},
representing \textbf{95.76\% of the retained frames telemetry corpus}.
This rigorous audit ensures that all subsequent statistical,
dimensionality reduction, clustering, and machine learning models
operate on true, synchronized physical observations.
\par}

\section{Related Work}\label{sec:related}

The processing and analysis of long-term Internet of Things (IoT) data
streams lie at the intersection of streaming anomaly detection,
unsupervised subspace learning, spatial clustering, and supervised drift
monitoring. This section reviews the theoretical and empirical
literature foundational to our pipeline architecture.

\subsection{Anomaly Detection in Edge Telemetry \& Sensor
Networks}\label{sub:rw_anomaly}
Anomaly detection in streaming sensor networks has been extensively
surveyed by \cite{chandola2009anomaly} and \cite{aggarwal2017outlier}. Traditional point-based methods rely on statistical
distributional assumptions, such as the univariate Z-score. However, as
noted by \cite{iglewicz1993detect}, standard Z-scores based on
sample mean and standard deviation break down in the presence of extreme
spikes due to their low breakdown point; robust estimators like the
Median Absolute Deviation (MAD) are required for non-Gaussian IoT
telemetry.

For multivariate streams, \cite{liu2008isolation} introduced the
\emph{Isolation Forest} algorithm, which isolates anomalies by randomly
partitioning feature space using axis-aligned trees. Unlike distance or
density estimators, Isolation Forest exhibits near-linear time complexity
\(O(n \log n)\), making it particularly effective for high-volume edge
streams. More recently, \cite{bhatia2022real} emphasized the
necessity of real-time edge processing architectures capable of
detecting anomalies under memory and bandwidth constraints, reinforcing
the importance of low-overhead C++ implementations over heavy Python
runtime environments.

\subsection{Subspace Methods \& Dimensionality Reduction}\label{sub:rw_subspace}
Principal Component Analysis (PCA), established by \cite{pearson1901lines} and generalized by \cite{hotelling1933analysis}, remains the
standard technique for modeling global linear dependencies among
continuous physical variables. In multivariate environmental sensing,
PCA transforms correlated measurements into orthogonal variance
components.

When deploying linear algebra algorithms directly on constrained IoT
gateways or embedded systems, external libraries (e.g., LAPACK,
Armadillo) introduce overhead and portability barriers. Golub and Van Loan \cite{golub2013matrix} outline numerical diagonalization algorithms,
specifically the Jacobi rotation method for symmetric matrices. For
bivariate correlation matrices (\(2 \times 2\)), closed-form algebraic
solutions bypass iterative numerical solvers entirely, providing exact
eigenvalue decomposition with zero iterative computational overhead.

\subsection{Partitioning and Density-Based Clustering}\label{sub:rw_clustering}

Cluster analysis serves a dual purpose in environmental telemetry:
segmenting distinct climatic regimes and isolating unstructured spatial
noise.

\begin{itemize}
\item
  \textbf{Partitioning Methods:} The classical K-means algorithm
  \cite{lloyd1982least} minimizes intra-cluster variance but is
  notoriously sensitive to initial centroid placement. \cite{arthur2007kmeans} addressed this with \emph{K-means++},
  introducing probabilistic distance-weighted initialization that
  guarantees an \(O(\log K)\) approximation bound. To determine the
  optimal number of clusters \(K\), \cite{rousseeuw1987silhouettes} proposed
  the \emph{silhouette index}, which quantifies cluster cohesion against
  separation.
\item
  \textbf{Density-Based Methods:} Unlike K-means, \textbf{DBSCAN}
  \cite{ester1996density} discovers clusters of arbitrary shapes
  and explicitly identifies noise points (labeled \(-1\)). However,
  naive DBSCAN scales at \(O(n^2)\) complexity and suffers from memory
  allocation limits (\texttt{std::bad\_alloc}) when expanding dense
  spatial queues on large datasets (\(n > 200{,}000\)). Mitigating this
  requires spatial indexing structures---such as uniform grid bins or
  \(k\)-d trees---and objective hyperparameter calibration via
  \(k\)-distance plot analysis.
\end{itemize}

\subsection{Predictive Modeling \& Drift Identification}\label{sub:rw_predict}

Supervised neural networks, specifically Multilayer Perceptrons (MLP)
trained via backpropagation \cite{rumelhart1986learning}, are
widely used to approximate nonlinear physical couplings between
environmental variables. In training deep or recurrent networks,
\cite{pascanu2013difficulty} demonstrated that \emph{gradient
clipping} and momentum-based stochastic gradient descent (SGD) stabilize
convergence against exploding gradients.

While point-based univariate and multivariate detectors (Z-score,
Isolation Forest, DBSCAN) evaluate each telemetry frame independently,
they cannot identify progressive calibration loss or sensor degradation
over multi-month horizons. As highlighted by \cite{aggarwal2017outlier},
cross-validating a physical predictive model (e.g., predicting relative
humidity from temperature and cyclical time encodings) allows large
regression residuals
(\(\vert{}y_i - \hat{y}_i\vert{} > 3 \times \mathrm{RMSE}\)) to serve as
a complementary indicator for slow, systemic sensor drifts that remain
within static physical bounds.

\subsection{Positioning Relative to Existing Multi-Method Pipelines
and Data-Quality Auditing
Practice}\label{positioning-relative-to-existing-multi-method-pipelines-and-data-quality-auditing-practice-1}

Sections~\ref{sub:rw_anomaly}--\ref{sub:rw_predict} each address anomaly detection, subspace projection,
clustering, or drift monitoring as largely separate bodies of
literature. Multi-method pipelines that \emph{combine} several of these
detectors into a single, complementary framework specifically for
LPWAN/IoT telemetry remain comparatively uncommon: most cited works
(e.g., \cite{liu2008isolation}; \cite{ester1996density}) evaluate
one detection paradigm in isolation, and surveys of IoT sensor-network
data quality tend to treat missing-data handling, outlier detection, and
drift correction as distinct tasks addressed by disjoint bodies of work
rather than a single orthogonal, four-tier architecture of the kind
proposed here. \cite{ferrercid2025review}, in a recent review of data-quality applications for IoT
monitoring sensor networks, make a closely related observation: despite
the wide range of individual techniques available for outlier detection,
missing-data imputation, and virtual sensing in environmental and
air-quality monitoring networks, no framework in their survey unifies
coverage auditing with cross-validated, complementary anomaly detection
at the pipeline level --- most deployments instead apply a single
technique per data-quality dimension. This gap is precisely what the
coverage-audit-first design proposed in Section~\ref{sub:acq}, combined with the
four-tier detection framework of Section~\ref{sub:math}, is intended to address: rather
than treating variable-coverage verification as a preliminary,
disposable step, this work integrates it directly into the extraction
pipeline and demonstrates empirically (Table~\ref{tab:coverage_audit}) that skipping it would
have silently invalidated the entire downstream analysis. To our
knowledge, no prior work specifically targets \textbf{LoRaWAN}
deployments with an integrated coverage-audit-and-multi-tier-detection
pipeline of this kind; positioning this contribution against the broader
(but still comparatively sparse) IoT data-quality-auditing literature,
rather than solely against single-technique anomaly detectors, is left
as recommended future work for a revised version of this paper.

\section{Theoretical Framework \& Methodology}\label{sec:theory}

\subsection{The Need for Multi-Tiered Anomaly Detection in IoT
Streams}\label{sub:multitier}
Anomalies in environmental time-series data do not manifest as a single
homogeneous class. In edge-computing and IoT sensor contexts, anomalies
span a spectrum of behaviors:

\begin{itemize}[leftmargin=*, label=\textbullet]
\item
  \textbf{Extreme Univariate Outliers:} Sensor spikes, electrical
  bursts, or out-of-spec hardware failures.
\item
  \textbf{Multivariate Structural Anomalies:} Observations where each
  individual feature remains within normal seasonal bounds, but their
  \emph{joint combination} violates fundamental physical laws or
  dominant environmental correlations.
\item
  \textbf{Density-Based Clustered Anomalies / Noise:} Isolated noise
  points residing in low-density regions of the multivariate feature
  space.
\item
  \textbf{Systemic Temporal Drifts:} Gradual, continuous calibration
  loss (e.g., aging capacitive humidity elements) that evades
  point-based detection because individual readings shift slowly over
  weeks or months.
\end{itemize}

No single detection algorithm can capture all four anomaly types
simultaneously. Therefore, this work establishes a \textbf{four-tier
complementary detection framework}, combining univariate statistics,
random partitioning, geometric subspace projection, and supervised
cross-validation.

\begin{figure*}[t]
\centering
\resizebox{0.9\textwidth}{!}{%
\begin{tikzpicture}[
  root/.style={draw, rounded corners=2pt, fill=boxfillblue, draw=accentblue,
               minimum width=6.4cm, minimum height=1cm, align=center, font=\small\bfseries, line width=0.9pt},
  tier/.style={draw, rounded corners=2pt, fill=boxfillgreen, draw=accentgreen,
               minimum width=3.5cm, minimum height=1.5cm, align=center, font=\scriptsize\bfseries, line width=0.8pt},
  note/.style={draw, rounded corners=2pt, fill=gray!6, draw=gray!55,
               minimum width=3.5cm, minimum height=1.7cm, align=center, font=\scriptsize, line width=0.6pt, text width=3.2cm},
  arr/.style={-{Latex[length=2mm]}, line width=0.8pt, draw=gray!60!black}
]
\node[root] (root) at (0,0) {\textbf{Clean Bivariate Dataset (T, H)}\\ $n = 205{,}005$};
\node[tier] (t1) at (-6.0,-2.6) {TIER 1(a)\\ Modified Z-Score\\ (Univariate)};
\node[tier] (t2) at (-2.0,-2.6) {TIER 1(b)\\ Isolation Forest\\ (Multivariate)};
\node[tier] (t3) at (2.0,-2.6)  {TIER 2\\ Bivariate PCA\\ Reconstruction\\ (Subspace 2D)};
\node[tier] (t4) at (6.0,-2.6)  {TIER 3\\ Cross-Val MLP\\ Residuals\\ (Temporal/Drift)};
\node[note] (n1) at (-6.0,-5.3) {Identifies extreme single-variable bounds ($\vert M_i\vert>3.5$)};
\node[note] (n2) at (-2.0,-5.3) {Identifies abnormal joint feature combinations ($s>0.62$)};
\node[note] (n3) at (2.0,-5.3)  {Quantifies departure from dominant T-H anti-correlation ($e_i = \vert y_{\mathrm{PC2}}\vert$)};
\node[note] (n4) at (6.0,-5.3)  {Identifies progressive calibration sensor drift ($\vert$res$\vert>3\times$RMSE)};
\draw[arr] (root.south) -- ++(0,-3mm) -| (t1.north);
\draw[arr] (root.south) -- ++(0,-3mm) -| (t2.north);
\draw[arr] (root.south) -- ++(0,-3mm) -| (t3.north);
\draw[arr] (root.south) -- ++(0,-3mm) -| (t4.north);
\draw[arr] (t1) -- (n1);
\draw[arr] (t2) -- (n2);
\draw[arr] (t3) -- (n3);
\draw[arr] (t4) -- (n4);
\end{tikzpicture}%
}
\caption{Four-tier complementary detection framework: each tier targets a distinct, orthogonal anomaly signature within the clean bivariate (T,H) dataset.}
\label{fig:cleanBivariate}
\end{figure*}
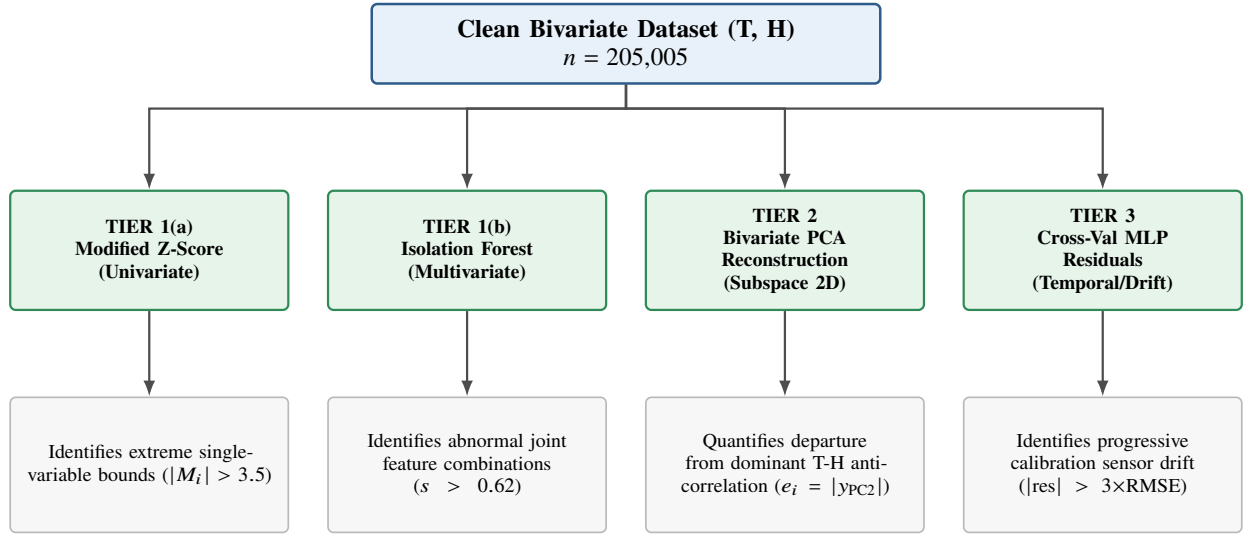

\subsection{Mathematical Formulation of the Complementary
Framework}\label{sub:math}
\subsubsection{Tier 1(a): Point-Based Univariate Detection --- Modified
Z-Score}

Classical \(Z\)-scores rely on the sample mean (\(\bar{x}\)) and
standard deviation (\(\sigma\)), both of which are notoriously sensitive
to extreme outliers (low breakdown point). To ensure robust estimation
over multi-year climate cycles, we employ the \textbf{Modified Z-Score}
based on the Median Absolute Deviation (MAD) \cite{iglewicz1993detect}:

\begin{equation}
\label{eq:modz}
M_i = 0.6745 \cdot \frac{x_i - \tilde{x}}{\mathrm{MAD}}
\end{equation}

where \(\tilde{x} = \mathrm{median}(x)\) and
\(\mathrm{MAD} = \mathrm{median}_i(\vert{}x_i - \tilde{x}\vert{})\). The
constant \(0.6745\) equates the MAD to the standard deviation under a
normal distribution. A point is flagged as a marginal univariate outlier
if \(\vert{}M_i\vert{} > 3.5\).

\subsubsection{Tier 1(b): Point-Based Multivariate Isolation ---Isolation Forest}

Univariate statistics cannot detect anomalous feature couplings. The
\textbf{Isolation Forest} algorithm \cite{liu2008isolation} exploits
the principle that anomalous observations are few and different, making
them susceptible to isolation via random recursive axis-aligned
hyperplanes.

An ensemble of \(t\) isolation trees partitions the standardized 2D
space \((T, H)\). The anomaly score \(s(x, n)\) for an observation \(x\)
over a dataset of size \(n\) is defined as:

\begin{equation}
\label{eq:iso_score}
s(x, n) = 2^{-\frac{E[h(x)]}{c(n)}}
\end{equation}

where \(E[h(x)]\) is the average path length across all isolation trees,
and \(c(n)\) is the average path length of unsuccessful searches in a
Binary Search Tree (BST):

%\[c(n) = 2H(n-1) - \frac{2(n-1)}{n}, \quad H(k) = \ln(k) + \gamma \quad (\gamma \approx 0.5772156649) \tag{3}\]
\begin{equation}
\label{eq:harmonic_cost}
\begin{multlined}
c(n) = 2H(n-1) - \frac{2(n-1)}{n}, \\
\quad H(k) = \ln(k) + \gamma \quad (\gamma \approx 0.5772)
\end{multlined}
\end{equation}

As \(s(x, n) \to 1\), the path length \(h(x) \to 0\), indicating an
easily isolated multivariate anomaly. We enforce an empirical isolation
threshold of \(s(x, n) > 0.62\).

\subsubsection{Tier 2: Geometric Subspace Anomaly Detection ---
Bivariate PCA Reconstruction
Error}

Principal Component Analysis (PCA) models the global covariance
structure. In a standardized 2D space (\(X'\)), the covariance matrix
\(S = \frac{1}{n-1}X'^\top X'\) is diagonalized to yield eigenvectors
\(V_1, V_2\) and eigenvalues \(\lambda_1 \ge \lambda_2\).

Retaining only the primary component (\(\mathrm{PC1}\), representing the
dominant thermo-hygrometric opposition axis), the projection matrix is
\(V_1\). The multivariate reconstruction error \(e_i\) is given by:

\begin{equation}
\label{eq:recon_error}
e_i = \big\lVert x'_i - V_1 V_1^\top x'_i \big\rVert_2
\end{equation}

In a 2D orthonormal basis, this reduces to the absolute magnitude of the
rejected second principal component:
\(e_i = \vert{}y_{i, \mathrm{PC2}}\vert{}\). Observations with high
\(e_i\) represent severe departures from the physical correlation
structure of the atmosphere (e.g., simultaneous high temperature and
high humidity).

\subsubsection{Tier 3: Temporal \& Drift Anomaly Detection ---
Supervised MLP
Cross-Validation}

Point-based methods (Modified Z-score, Isolation Forest, PCA) evaluate
each frame as an isolated tuple, remaining blind to \textbf{slow sensor
drifts} where values drift gradually while staying inside valid static
bounds.

To capture temporal inconsistencies, a Multilayer Perceptron (MLP) is
trained to predict \(H\) using Temperature \(T\) alongside cyclically
encoded temporal features:

%\[\mathbf{x}_{\mathrm{time}} = \left[ \sin\left(\frac{2\pi \cdot \text{hour}}{24}\right), \cos\left(\frac{2\pi \cdot \text{hour}}{24}\right), \sin\left(\frac{2\pi \cdot \text{day}}{365.25}\right), \cos\left(\frac{2\pi \cdot \text{day}}{365.25}\right) \right] \tag{5}\]

\begin{multline}
\label{eq:time_features}
\mathbf{x}_{\mathrm{time}} = \Big[ \sin(2\pi \cdot \text{hour} / 24), \, \cos(2\pi \cdot \text{hour} / 24), \\
\sin(2\pi \cdot \text{day} / 365.25), \, \cos(2\pi \cdot \text{day} / 365.25) \Big]
\end{multline}

The network architecture (\(16\) hidden units, \(\tanh\) activation)
minimizes the Mean Squared Error:

\begin{equation}
\label{eq:mse}
L = \frac{1}{n}\sum_{i=1}^n (y_i - \hat{y}_i)^2
\end{equation}

An observation \(i\) that is \textbf{not flagged} by Tier 1 or Tier 2,
yet exhibits a large prediction residual
\(\vert{}y_i - \hat{y}_i\vert{} > 3 \times \mathrm{RMSE}\), serves as a
prime indicator of localized sensor drift or unmodeled environmental
micro-disturbances.

\subsection{Methodological Orthogonality
Summary}\label{sub:orthogonality}
The theoretical necessity for this multi-tiered architecture is
summarized in the matrix below:
% >>> begin table_orthogonality.tex
\begin{table}[t]
\centering
\caption{Methodological orthogonality of the four detection tiers.}
\label{tab:orthogonality}
%\footnotesize
\scriptsize
\renewcommand{\arraystretch}{1.3}
\setlength{\tabcolsep}{3pt}
%\begin{tabularx}{\columnwidth}{@{} >{\raggedright\arraybackslash}p{2.5cm} X X X @{}}
\begin{tabularx}{\columnwidth}{@{} >{\raggedright\arraybackslash}p{2cm} L L L @{}}
\toprule
\textbf{Detection Tier} & \textbf{Underlying Principle} & \textbf{Primary Target} & \textbf{Insensitive To} \\
\midrule
Modified Z-Score & Marginal distance from median, in units of MAD & Extreme single-variable spikes ($T$ or $H$) & Multivariate correlation violations within normal marginal bounds \\
Isolation Forest & Random axis-aligned partitioning path length & Unusual joint combinations of $(T,H)$ & Slow, continuous calibration drifts over long horizons \\
PCA Reconstruction & Distance to the primary variance subspace & Departure from T--H anti-correlation ($e_i=\lvert y_{\mathrm{PC2}}\rvert$) & Anomalies aligned along the primary variance axis (PC1) \\
DBSCAN Noise Label & Local density threshold ($N_\varepsilon(p) < \mathrm{minPts}$) & Sparse spatial outliers, isolated noise points & Uniform dense clusters from sensor quantization \\
MLP Cross-Validation & Deviation from learned physical/temporal dependency & Slow sensor drift, temporal micro-climate anomalies & Sudden point anomalies uncoupled from diurnal/seasonal trends \\
\bottomrule
\end{tabularx}
\end{table}
% <<< end table_orthogonality.tex

\section{Data Processing \& Analytical Pipeline}\label{sec:pipeline}

\subsection{Data Acquisition, Coverage Audit \&
Cleaning}\label{sub:acq}
The primary data acquisition and preprocessing layer is implemented in
\texttt{01\_extract.cpp}. The raw corpus comprises \(421{,}937\)
JSON-Lines frames transmitted by five distinct LoRaWAN nodes deployed at
the Perret Tower (\emph{Tour Perret}) in Grenoble, France, between June
2021 and June 2023, decoded by a ChirpStack network server \cite{donsez2023tour} (DOI 10.18709/perscido.2023.06.ds395).

% >>> begin fig_pipeline_architecture.tex
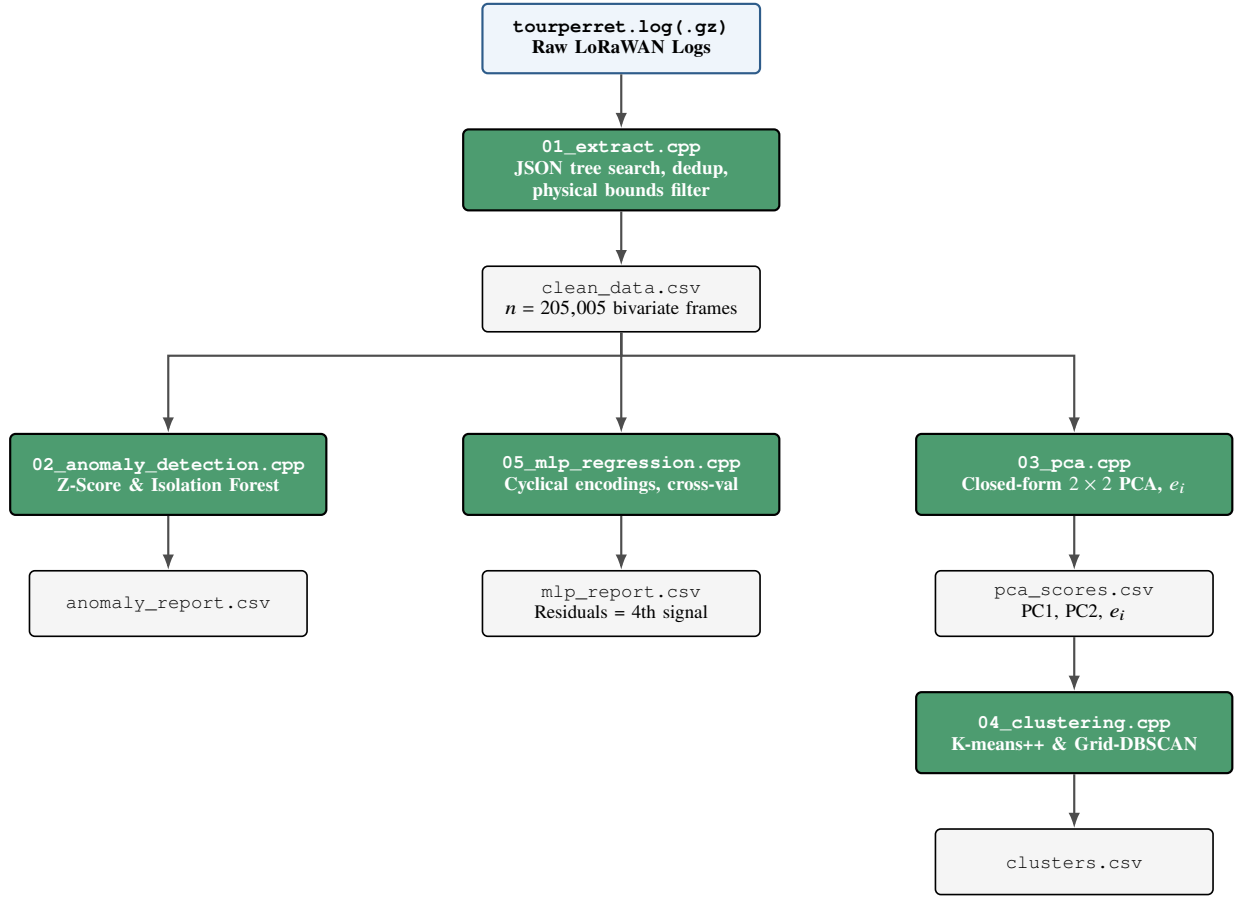
\begin{figure*}[t]
\centering
\resizebox{0.9\textwidth}{!}{%
\begin{tikzpicture}[
  node distance=7mm and 8mm,
  src/.style={draw, rounded corners=2pt, fill=boxfillblue!70, draw=accentblue, minimum width=3.6cm, minimum height=0.9cm, align=center, font=\scriptsize\bfseries, line width=0.8pt},
  mod/.style={draw, rounded corners=2pt, fill=accentgreen!85, text=white, minimum width=4.1cm, minimum height=1.05cm, align=center, font=\scriptsize\bfseries, line width=0.8pt},
  outbox/.style={draw, rounded corners=2pt, fill=gray!8, minimum width=3.6cm, minimum height=0.85cm, align=center, font=\scriptsize, line width=0.6pt},
  arr/.style={-{Latex[length=2mm]}, line width=0.8pt, draw=gray!60!black}
]
\node[src] (raw) {\texttt{tourperret.log(.gz)}\\ Raw LoRaWAN Logs};
\node[mod, below=of raw] (ext) {\texttt{01\_extract.cpp}\\ JSON tree search, dedup,\\ physical bounds filter};
\node[outbox, below=of ext] (clean) {\texttt{clean\_data.csv}\\ $n=205{,}005$ bivariate frames};
\node[mod, below left=13mm and 20mm of clean] (anom) {\texttt{02\_anomaly\_detection.cpp}\\ Z-Score \& Isolation Forest};
\node[mod, below=13mm of clean] (mlp) {\texttt{05\_mlp\_regression.cpp}\\ Cyclical encodings, cross-val};
\node[mod, below right=13mm and 20mm of clean] (pca) {\texttt{03\_pca.cpp}\\ Closed-form $2\times2$ PCA, $e_i$};
\node[outbox, below=of anom] (anomout) {\texttt{anomaly\_report.csv}};
\node[outbox, below=of mlp] (mlpout) {\texttt{mlp\_report.csv}\\ Residuals = 4th signal};
\node[outbox, below=of pca] (pcaout) {\texttt{pca\_scores.csv}\\ PC1, PC2, $e_i$};
\node[mod, below=of pcaout] (clus) {\texttt{04\_clustering.cpp}\\ K-means++ \& Grid-DBSCAN};
\node[outbox, below=of clus] (clusout) {\texttt{clusters.csv}};
\draw[arr] (raw) -- (ext);
\draw[arr] (ext) -- (clean);
\draw[arr] (clean.south) -- ++(0,-3mm) -| (anom.north);
\draw[arr] (clean) -- (mlp);
\draw[arr] (clean.south) -- ++(0,-3mm) -| (pca.north);
\draw[arr] (anom) -- (anomout);
\draw[arr] (mlp) -- (mlpout);
\draw[arr] (pca) -- (pcaout);
\draw[arr] (pcaout) -- (clus);
\draw[arr] (clus) -- (clusout);
\end{tikzpicture}
}
\caption{Data processing and analytical pipeline: from raw LoRaWAN logs to the four output reports (anomaly, MLP, PCA/cluster) that feed the Results section.}
\label{fig:pipeline_architecture}
\end{figure*}
% <<< end fig_pipeline_architecture.tex

\subsubsection{JSON Schema Robustness \& Coverage
Audit}

Rather than querying rigid, static path keys, \texttt{01\_extract.cpp}
executes a recursive, alias-tolerant search across each frame's JSON
payload hierarchy (\texttt{object.*} for physical measurements,
\texttt{rxInfo{[}{]}} for reception metadata). This handles schema
variations across ChirpStack v3/v4, The Things Network (TTN) v3, and
flat-format payloads without throwing runtime key-errors. Two
schema-level pitfalls, identified by diagnostic inspection of the raw
file prior to full extraction, required specific handling: (i) the
timestamp field is neither \texttt{time} nor \texttt{timestamp} but
appears as either \texttt{\_date} (an ISO 8601 string) or
\texttt{\_timestamp} (a \textbf{Unix epoch in milliseconds}, not seconds
--- a common source of silent date-arithmetic errors if left
unconverted); and (ii) Elsys EMS frames in this deployment do
\textbf{not} always carry all four target variables jointly --- the
first frames inspected during development carried only
\texttt{temperature} and \texttt{humidity} (alongside unrelated fields
such as \texttt{dewpoint}, \texttt{accMotion}, \texttt{pulseAbs},
\texttt{vdd}, \texttt{waterleak}, \texttt{x/y/z}). An early extraction
attempt requiring all four variables in the same frame therefore
rejected the \emph{entire} corpus, including frames where \(T\) and
\(H\) were both perfectly valid --- motivating the per-variable
extraction strategy below.

Each of the four target variables (\(T\), \(H\), \(P\),
\(\mathrm{CO_2}\)) is consequently extracted \textbf{independently}: a
frame is retained as soon as it carries \emph{at least one} target
variable and a valid timestamp, with absent columns left empty in a
sparse (``long'') intermediate CSV rather than requiring a strict
wide-format join. This design choice is what makes the coverage audit
below possible in the first place --- it lets per-variable and
per-\texttt{devEUI} presence be measured across the \emph{entire} corpus
without first discarding rows for missing columns.

During extraction, an automated \textbf{Data Coverage Audit} evaluates
payload completeness across all registered device identifiers
(\texttt{devEUI}) and application boundaries (\texttt{applicationName})
in a single pass.

% >>> begin table2.tex
\begin{table}[htbp]
\centering
\caption{Coverage per variable and per device (audit on the 421{,}937 raw frames).}
\label{tab:coverage_audit}
\footnotesize
\setlength{\tabcolsep}{2pt}
\begin{tabularx}{\columnwidth}{@{} >{\raggedright\arraybackslash}p{1.6cm} r r r r L c @{}}
\toprule
\textbf{Variable} & \textbf{Raw} & \textbf{Retained} & \textbf{Cov.} & \textbf{Cov.} & \textbf{Scope} & \textbf{Nodes} \\
 & \textbf{Frames} & & \textbf{(\%Tot.)} & \textbf{(\%Ret.)} & & \\
\midrule
Temperature ($T$)       & 393,685 & 214,076            & 93.30 & 100.00 & Elsys + Wyres & 5 \\
Humidity ($H$)          & 384,614 & 205,005            & 91.15 & 95.76 & Elsys only    & 4 \\
Pressure ($P$)          & 9,071   & 0 \emph{(isolated)}  & 2.15 & 0.00 & Wyres only    & 1 \\
Carbon Dioxide (CO$_2$) & 0         & 0 \emph{(absent)}    & 0.00 & 0.00 & ---           & 0 \\
\bottomrule
\end{tabularx}
\end{table}
% <<< end table2.tex

The coverage audit confirms, on the full real-world corpus, what the
early four-variable extraction attempt had already suggested
empirically: a static extraction assuming joint co-occurrence across all
four initially planned variables (\(T, H, P, \mathrm{CO_2}\)) yields
\textbf{zero rows} by construction (complete-case filtering on an empty
intersection). This audit --- integrated into the extractor and executed
in a single pass over the file --- is itself the methodological step
whose necessity this work argues for: without it, the four-variable
hypothesis would have failed silently rather than being diagnosed and
corrected.

\textbf{Distribution by \texttt{devEUI} (214,076 retained frames):} 4
Elsys EMS sensors (108,678, 35,387, 30,800, and 30,140 frames
respectively, totaling 205,005 joint T+H frames) and 1 Wyres module (9,071
T+P frames, no humidity).

\subsubsection{Multi-Gateway Deduplication \& Physical Bounds
Filtering}

In LoRaWAN deployments, a single wireless transmission frame is
frequently picked up by multiple surrounding gateways. Frame
deduplication is performed using a composite hash key combining the
unique device identifier and the frame counter:

\begin{equation}
\label{eq:dedup}
\text{Deduplication Key} = \mathtt{devEui} \parallel \mathtt{fCount}
\end{equation}

Deduplication eliminated \(179{,}609\) duplicate frame receipts
(\(42.6\%\) of total ingested logs), reflecting the dense gateway
coverage across the Grenoble urban area. Following deduplication,
physical boundary validation was applied based on the Elsys EMS
manufacturer specification sheet~\cite{elsys_ems_spec}:

\begin{itemize}
\item
  \textbf{Temperature:} \(T \in [-40.0, 85.0]\text{\textdegree C}\)
\item
  \textbf{Relative Humidity:} \(H \in [0.0, 100.0]\%\)
\end{itemize}

Only three individual out-of-range sensor readings were discarded. The
resulting retained core dataset consists of \(n = 205{,}005\)
synchronized, bivariate \((T, H)\) frames from four Elsys EMS sensors
(\(108{,}678\), \(35{,}387\), \(30{,}800\), and \(30{,}140\) frames
respectively), representing \textbf{\(95.76\%\) of the retained
deduplicated corpus}.

\subsection{Anomaly Detection
Framework}\label{sub:anomaly}

The C++ module \texttt{02\_anomaly\_detection.cpp} implements the three
point/density-based tiers of the complementary detection framework
introduced in Section~\ref{sub:math}, refocused on the bivariate \((T, H)\) space jointly
available across the \(205{,}005\) Elsys EMS frames --- down from the
four dimensions initially planned.

\subsubsection{(a) Point-Based Univariate Outliers: Robust Modified
Z-Score}

The Modified Z-score (Eq.~\eqref{eq:modz}, Section~\ref{sub:math}) is computed independently for \(T\)
and \(H\); observations with \(\vert{}M_i\vert{} > 3.5\) are flagged as
univariate marginal outliers.

\subsubsection{(b) Multivariate Point Outliers: Isolation
Forest}\label{b-multivariate-point-outliers-isolation-forest-1}

The Isolation Forest anomaly score (Eqs.~\eqref{eq:iso_score}--\eqref{eq:harmonic_cost}, Section~\ref{sub:math}) is evaluated on
standardized bivariate vectors \(\mathbf{x}_i = [T'_i, H'_i]^\top\). The
forest is instantiated with \(t = 100\) isolation trees, each built on a
subsample of \(\psi = 256\) observations, with maximum tree depth capped
at \(h_{\mathrm{max}} = \lceil \log_2(256) \rceil = 8\). As in Section~\ref{sub:math}, an
empirical decision threshold of \(s(\mathbf{x}, n) > 0.62\) is enforced.

\subsubsection{(c) Global multivariate structure:
DBSCAN}

\textbf{DBSCAN} (\texttt{04\_clustering.cpp}, Section~\ref{sub:clustering} and Section~\ref{sub:res_cluster}) contributes
a third, orthogonal detector: points located in low-density regions of
the PCA space (label \(-1\)) reflect the global multivariate structure
rather than a per-variable score or random-tree isolation path.

\subsection{Dimensionality Reduction: Closed-Form Bivariate
PCA}\label{sub:pca}\label{dimensionality-reduction-closed-form-bivariate-pca-1}

The module \texttt{03\_pca.cpp} performs the Tier-2 subspace projection
introduced in Section~\ref{sub:math} (reconstruction error \(e_i\), Eq.~\eqref{eq:recon_error}), and derives
its exact closed-form solution for the bivariate case.

\subsubsection{Standardization \& Closed-Form
\texorpdfstring{$2 \times 2$}{2 x 2} Diagonalization}

Given raw features with incompatible physical units (\textdegree C and
\(\%\)), features are standardized to zero mean and unit variance:
\(x'_{ij} = (x_{ij} - \bar{x}_j) / \sigma_j\). The \(2 \times 2\) sample
correlation matrix \(S\) is expressed as:

\begin{equation}
\label{eq:corr}
S = \frac{1}{n-1} X'^\top X' = \begin{bmatrix} 1 & r_{T,H} \\ r_{T,H} & 1 \end{bmatrix}
\end{equation}

While general multivariate PCA (\(d \ge 3\)) requires numerical solvers
such as the Jacobi rotation algorithm \cite{golub2013matrix}, a
\(2 \times 2\) symmetric matrix admits an exact closed-form algebraic
solution for its eigenvalues \(\lambda_1, \lambda_2\):

\begin{equation}
\label{eq:eigen}
\lambda_{1,2} = \frac{\mathrm{tr}(S) \pm \sqrt{\mathrm{tr}(S)^2 - 4\det(S)}}{2} = 1 \pm \vert{}r_{T,H}\vert{}
\end{equation}

The corresponding orthonormal eigenvectors
\(\mathbf{v}_1, \mathbf{v}_2\) define the principal subspace projection
matrix \(V = [\mathbf{v}_1 \mid \mathbf{v}_2]\):

\begin{equation}
\label{eq:eigvec}
\mathbf{v}_1 = \frac{1}{\sqrt{2}} \begin{bmatrix} \mathrm{sgn}(r_{T,H}) \\ 1 \end{bmatrix}, \quad \mathbf{v}_2 = \frac{1}{\sqrt{2}} \begin{bmatrix} -\mathrm{sgn}(r_{T,H}) \\ 1 \end{bmatrix}
\end{equation}

Substituting \(\mathbf{v}_1\) into the reconstruction error of Eq.~\eqref{eq:recon_error}
(Section~\ref{sub:math}) and exploiting the 2D orthonormal basis, \(e_i\) reduces directly
to the absolute magnitude of the rejected PC2 coordinate,
\(e_i = \vert{}y_{i, \mathrm{PC2}}\vert{}\) --- this closed form is what
\texttt{03\_pca.cpp} evaluates for every observation. Observations with
high \(e_i\) represent severe departures from the dominant physical
correlation structure of the atmosphere.

\subsection{Clustering Algorithms \& Real-World Scaling
Optimizations}\label{sub:clustering}

The module \texttt{04\_clustering.cpp} implements K-means++ and
DBSCAN for spatial partitioning and density analysis.

\subsubsection{K-means++ Inertia Minimization \& Silhouette
Optimization}

K-means minimizes intra-cluster variance (inertia) \(J\):

\begin{equation}
\label{eq:kmeans}
J = \sum_{k=1}^K \sum_{\mathbf{x} \in C_k} \left\Vert{} \mathbf{x} - \boldsymbol{\mu}_k \right\Vert{}^2
\end{equation}

Using Lloyd's algorithm \cite{lloyd1982least} with K-means++ seeding \cite{arthur2007kmeans}, initial cluster centers are sampled with
probability proportional to their squared distance to the nearest
existing center:
\(P(\mathbf{x}) = \frac{D(\mathbf{x})^2}{\sum D(\mathbf{x})^2}\). To
select the optimal number of clusters \(K\), the silhouette index
\(s(i)\) \cite{rousseeuw1987silhouettes} is computed across candidate evaluations
\(K \in \{2, 3, \dots, 8\}\):

\begin{equation}
\label{eq:silhouette}
s(i) = \frac{b(i) - a(i)}{\max(a(i), b(i))}
\end{equation}

where \(a(i)\) represents mean intra-cluster distance and \(b(i)\)
denotes mean nearest-neighbor cluster distance.

\subsubsection{DBSCAN Spatial Indexing \& Algorithmic
Scaling \texorpdfstring{(\(n = 205{,}005\))}{(n = 205,005)}}

DBSCAN \cite{ester1996density} defines clusters as continuous high-density
regions separated by low-density space. A point \(\mathbf{p}\) is a core
point if its \(\varepsilon\)-neighborhood \(N_\varepsilon(\mathbf{p})\)
contains at least \(\mathrm{minPts}\) points:

\begin{multline}
\label{eq:dbscan_nbhd}
N_\varepsilon(\mathbf{p}) = \{\,\mathbf{q} \in \mathcal{D} \mid \Vert\mathbf{p}-\mathbf{q}\Vert \le \varepsilon\,\}, \\
\text{subject to } \vert N_\varepsilon(\mathbf{p})\vert \ge \mathrm{minPts}
\end{multline}

Points not density-reachable from any core point are flagged as noise
(label \(-1\)).

% --- FIGURE BLOCK START ---
\begin{figure}[H]
  \centering
  \includegraphics[width=1\columnwidth]{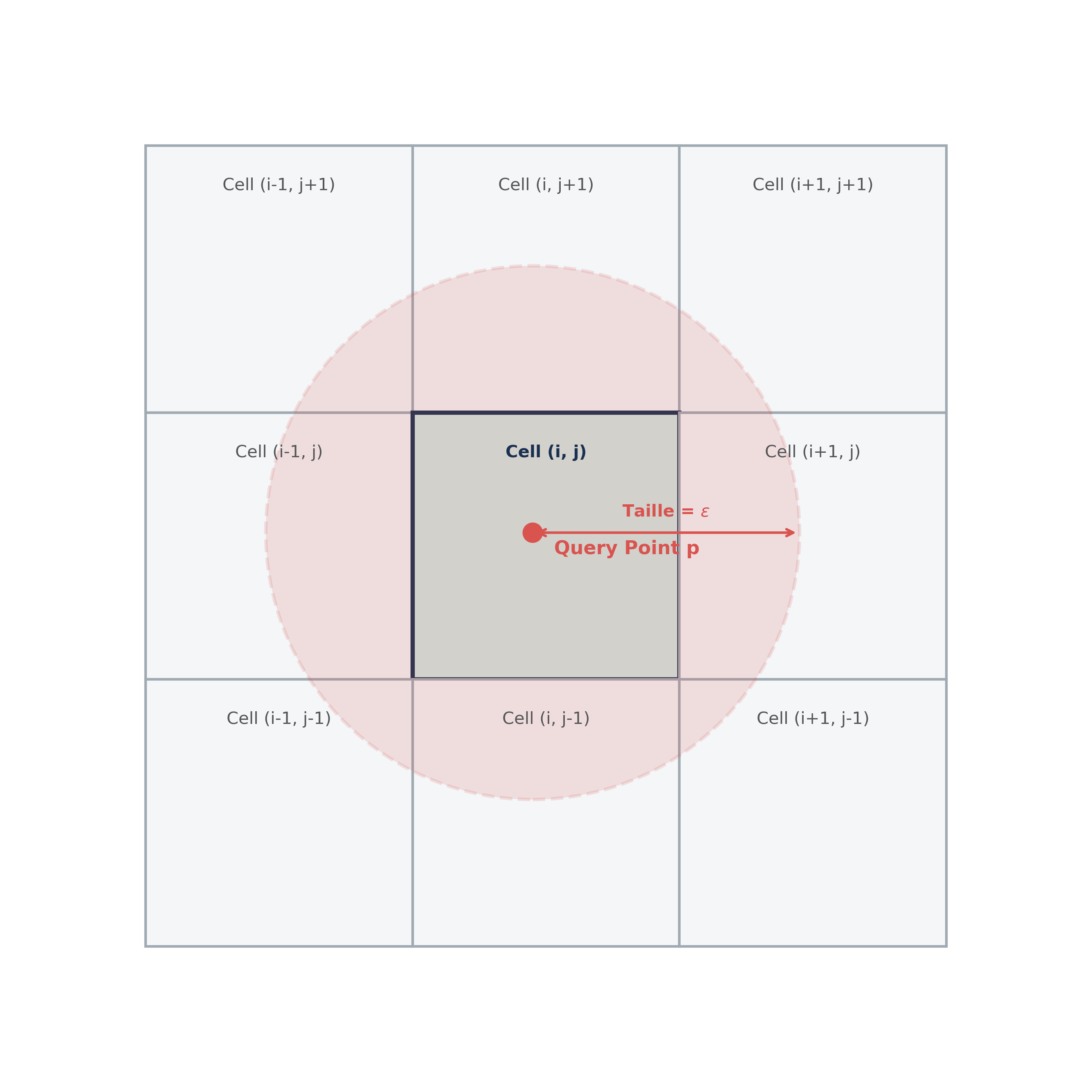}
  \caption{Uniform grid spatial index (2D), cell size $=\varepsilon$: DBSCAN's $\varepsilon$-neighborhood query is restricted to a $3\times3$ neighborhood of grid cells around the query point's cell, avoiding an $O(n^2)$ full scan.}
  \label{fig:spatial_grid_index-wt}
\end{figure}
% --- FIGURE BLOCK END ---

\subsubsection{Real-World Scaling
Challenges}

Executing naive \(O(n^2)\) spatial range queries on \(n = 205{,}005\)
bivariate points leads to two computational failures: excessive
execution times and memory allocation faults (\texttt{std::bad\_alloc})
caused by unchecked expansion queues in high-density regions. To resolve
these limitations, \texttt{04\_clustering.cpp} incorporates two
structural optimizations:

\begin{enumerate}
\def\labelenumi{\arabic{enumi}.}
\item
  \textbf{Uniform Grid Spatial Indexing:} Standardized PCA feature space
  is partitioned into a uniform 2D grid with square bin width equal to
  \(\varepsilon\). Range queries for point \(\mathbf{p} = (x, y)\) are
  restricted to candidate points residing within its home grid cell
  \((i, j)\) and the 8 immediate neighboring cells. Under quasi-uniform
  spatial distributions, search complexity scales from \(O(n^2)\) down
  to \(O(n)\).
\item
  \textbf{Expansion Queue State Tracking:} A boolean visited array
  (\texttt{std::vector\textless{}bool\textgreater{}\ in\_queue}) tracks
  point assignments during region expansion, preventing identical points
  from being repeatedly queued and eliminating runaway memory
  allocation.
\item
  \textbf{Hyperparameter Calibration via \(k\)-Distance Plot} Rather
  than using arbitrary default parameters (e.g.,
  \(\varepsilon = 0.35\)), parameter selection is driven by
  \(k\)-distance analysis with \(k = \mathrm{minPts} = 10\). For every
  point in the \(n = 205{,}005\) dataset, the Euclidean distance to its
  \(10\text{th}\)-nearest neighbor was calculated and sorted in
  ascending order. The maximum curvature point (``knee'') of the
  resulting curve occurs at \(\varepsilon \approx 0.071\). Setting
  \(\varepsilon = 0.35\) proved too coarse, merging \(100\%\) of points
  into a single cluster and returning zero noise points.
\end{enumerate}

% --- FIGURE BLOCK START ---
\begin{figure}[H]
  \centering
  \includegraphics[width=0.9\columnwidth]{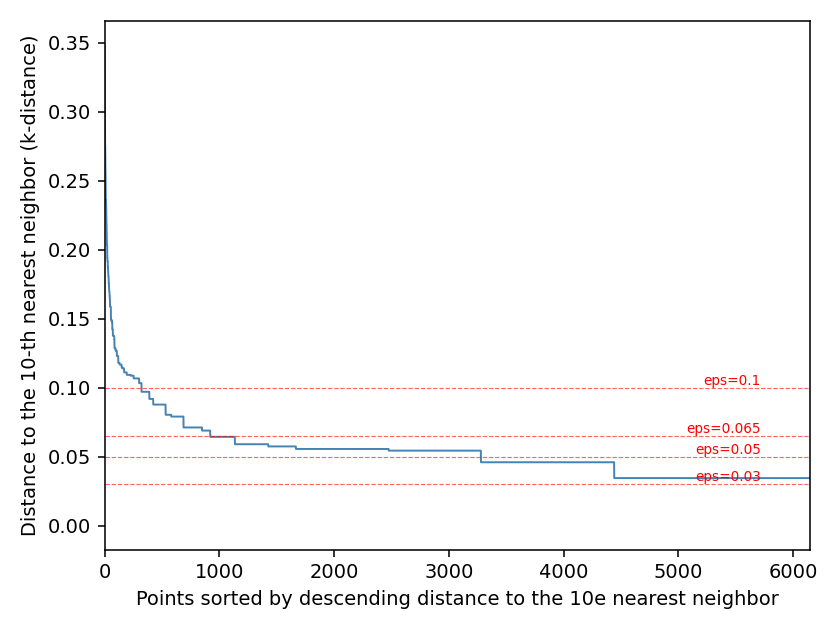}
  \caption{$k$-distance plot ($k=\mathrm{minPts}=10$) used to objectively select the DBSCAN neighborhood radius $\varepsilon$ on real-world data. Candidate thresholds are overlaid; the knee of the curve, identified via maximum perpendicular distance to the chord, is retained at $\varepsilon \approx 0.071$.}
  \label{fig:fig6_kdistance-en-wt}
\end{figure}
% --- FIGURE BLOCK END ---

\subsection{Cross-Validation via Supervised Learning
(MLP)}\label{sub:mlp}

The module \texttt{05\_mlp\_regression.cpp} implements a Multilayer
Perceptron (MLP) to model physical dependencies between variables and
detect progressive sensor drift.

\subsubsection{Target Revision \& Physical
Rationale}\label{target-revision-physical-rationale-1}

In the initial project design, the MLP was specified to predict
\(\mathrm{CO_2}\) concentration from thermal and pressure covariates.
However, the data coverage audit (Section~\ref{sub:acq}) proved that \(\mathrm{CO_2}\) is
entirely absent from the multi-year corpus (\(0\%\) presence).

To preserve the predictive cross-validation architecture, the target
variable was revised to predict \textbf{Relative Humidity (\(H\))} using
\textbf{Temperature (\(T\))} and temporal features. Relative Humidity is
especially prone to hardware degradation and capacitive element
saturation under prolonged humid conditions, making it an ideal target
for cross-validation drift analysis.

\subsubsection{Cyclical Feature Encoding \&
Architecture}

Time components are cyclically encoded exactly as defined in Eq.~\eqref{eq:time_features}
(Section~\ref{sub:math}), preventing artificial boundary discontinuities at daily
(\(23\mathrm{h} \to 00\mathrm{h}\)) and annual
(\(\text{Dec } 31 \to \text{Jan } 1\)) transitions. The input vector
\(\mathbf{x}_i = \left[ T'_i, \mathbf{x}_{\mathrm{time}} \right]^\top \in \mathbb{R}^5\)
feeds a feedforward network comprising one hidden layer with 16 neurons
(\(\tanh\) activation) and a single linear output unit predicting
\(\hat{H}_i\).

Network parameters are optimized using stochastic gradient descent (SGD)
with momentum (\(\mu = 0.9\), learning rate \(\eta = 0.002\)) and
gradient norm clipping \cite{pascanu2013difficulty}, minimizing the MSE loss
of Eq.~\eqref{eq:mse} (Section~\ref{sub:math}):

\begin{equation}
\label{eq:sgd}
\mathbf{v}_{t+1} = \mu \mathbf{v}_t - \eta \nabla L, \quad W_{t+1} = W_t + \mathbf{v}_{t+1}
\end{equation}

\subsubsection{High Residuals as Long-Term Drift
Signals}

Point-based detectors evaluate individual frames as independent events,
making them blind to slow calibration drifts where readings remain
within static physical bounds. By cross-validating physical dependencies
through the neural network, observations exhibiting large regression
residuals:

\begin{equation}
\label{eq:drift}
\left\vert{} y_i - \hat{y}_i \right\vert{} > 3 \times \mathrm{RMSE}
\end{equation}

serve as an independent fourth anomaly indicator targeting progressive
sensor drift and micro-climatic disruptions.

\subsection{Synthetic Validation Set: Generation Procedure and Role
in the Study
Design}\label{sub:synthetic}

Before being applied to the real-world Tour Perret corpus, every module
of the pipeline (Sections~\ref{sub:anomaly}--\ref{sub:mlp}) was first exercised against a
\textbf{synthetic bivariate Temperature--Humidity dataset}
(\(n = 20{,}000\) samples) generated with a known, near-deterministic
thermo-hygrometric anti-correlation (target \(r \approx -0.93\)). This
synthetic set was not designed as a formal experimental baseline to be
reported as a standalone result; rather, it served as a
\textbf{pre-deployment calibration and software-verification harness},
distinct from the numerical cross-check already described for
\texttt{03\_pca.cpp} --- in which the C++ eigen-decomposition is
independently recomputed in Python (\texttt{numpy.linalg.eigh}) on the
\emph{real} data to confirm bit-for-bit agreement on
\(\mathrm{PC1} = 83.10\%\), \(\mathrm{PC2} = 16.90\%\) (Section~\ref{sub:pca}). The
synthetic set, by contrast, was used earlier in the pipeline's
development, before the idiosyncrasies of the real corpus (variable
sparsity, actual point density) were fully characterized. It played
three concrete roles:

\begin{enumerate}
\def\labelenumi{\arabic{enumi}.}
\item
  \textbf{Default-parameter calibration and its failure mode (Section~\ref{sub:clustering}).}
  DBSCAN's initial density threshold was calibrated on the synthetic set
  at \(\varepsilon = 0.35\). Applied unchanged to the real corpus
  (\(n = 205{,}005\), substantially denser in standardized PCA space than
  the synthetic sample), this value proved \textbf{far too permissive}:
  it collapsed the entire dataset into a single cluster with \(0\%\)
  noise --- an uninformative result that made clustering structure
  impossible to assess. This failure is itself an informative finding:
  it demonstrates that DBSCAN hyperparameters calibrated on a synthetic
  proxy do not transfer to the real deployment, and it directly
  motivated the data-driven \(k\)-distance recalibration described in
  Section~\ref{sub:clustering}, which yielded the retained value \(\varepsilon \approx 0.071\).
\item
  \textbf{MLP negative control (Section~\ref{sub:res_mlp}).} For the regression
  cross-validation, the synthetic humidity target was additionally
  corrupted with \textbf{deliberately strong multiplicative noise},
  calibrated so that no learnable \((T, \text{time}) \to H\)
  relationship would remain. The MLP trained on this corrupted synthetic
  target achieved \(R^2 = 0.019\) --- statistically indistinguishable
  from a non-informative predictor. This negative-control result is the
  reference against which the real-world performance (\(R^2 = 0.672\),
  Section~\ref{sub:res_mlp}) should be read: it shows that the architecture does not
  spuriously report a high \(R^2\) in the absence of genuine signal,
  which in turn supports treating the real-world \(R^2 = 0.672\) as
  evidence of a real, learnable thermo-hygrometric and diurnal/seasonal
  dependency rather than an artifact of model capacity or overfitting.
\item
  \textbf{Visualization scale reference (Figure~\ref{fig:fig4_pca_clusters-en-wt}).} The synthetic
  sample size (\(n = 20{,}000\)) is markedly smaller and less dense than
  the real corpus (\(n = 205{,}005\)); marker size and transparency in
  the PCA projection scatter plots were adjusted accordingly (smaller,
  more transparent markers) once the pipeline moved from the synthetic
  development set to the real, denser point cloud.
\end{enumerate}

\textbf{Interpretation.} Read together, the correlation and \(R^2\)
comparisons in Section~\ref{sub:res_pca} and Section~\ref{sub:res_mlp} are not a same-footing benchmark between
two datasets of equal experimental status: the synthetic set encodes a
clean, controllable ground truth used to verify that each detector
behaves as expected (or, in the case of DBSCAN's initial
\(\varepsilon\), to expose a calibration that does \emph{not} transfer)
before the same code is trusted on field data. The lower real-world
correlation magnitude (\(\vert r_{T,H}\vert = 0.662\) vs.~\(0.93\) on
the synthetic set) reflects genuine microclimatic variability, weather
events, and sensor noise absent from the synthetic generator (Section~\ref{sub:res_pca}); the
higher real-world \(R^2\) (0.672 vs.~0.019 on the noise-corrupted
synthetic set) confirms the MLP is picking up a genuine physical signal
rather than an artifact of the fitting procedure (Section~\ref{sub:res_mlp}).

%% ─────────────────────────────────────────────────────────────────────────────

\section{Results}\label{sec:results}

\subsection{Descriptive Statistics
\texorpdfstring{(\(n = 205{,}005\), Complete Elsys EMS T+H Frames)}{(n = 205,005, Complete Elsys EMS T+H Frames)}}\label{sub:desc}

% >>> begin table2.tex
\begin{table}[H]
\centering
\caption{Statistical summary of the $n=205{,}005$ synchronized Elsys EMS bivariate telemetry corpus.}
\label{tab:descriptive_stats}
\scriptsize
\setlength{\tabcolsep}{2pt}
\begin{tabularx}{\columnwidth}{@{} l *{7}{Y} @{}}
\toprule
\textbf{Variable} & \textbf{Mean} & \textbf{Std.\ Dev.} & \textbf{Min} & \textbf{Q1 (25\%)} & \textbf{Median} & \textbf{Q3 (75\%)} & \textbf{Max} \\
\midrule
Temperature $T$ (\textdegree C) & 14.36 & 8.72  & $-4.30$ & 7.00  & 14.80 & 20.70 & 46.10 \\
Humidity $H$ (\%)               & 72.14 & 18.41 & 19.00   & 58.00 & 74.00 & 88.00 & 100.00 \\
\bottomrule
\end{tabularx}
\end{table}
% <<< end table2.tex

\subsubsection{Physical Interpretation \& Measurement
Artifacts}

\begin{itemize}
\item
  \textbf{Thermal Extremes (\(T_{\mathrm{max}} = 46.10\text{°C}\)):} The
  maximum recorded temperature exceeds ambient climate norms for
  Grenoble. This reflects localized solar irradiance heating on the
  unshaded exterior housing of the rooftop node.
\item
  \textbf{Humidity Saturation (\(H_{\mathrm{max}} = 100.00\%\)):}
  Relative humidity exhibits strong left-skewness, with a third quartile
  of \(\mathrm{Q3} = 88.00\%\) and frequent saturation at \(100.00\%\).
  This behavior corresponds to capacitive sensor saturation during
  sustained rainfall or high atmospheric humidity.
\end{itemize}

\subsection{Bivariate PCA Analysis}\label{sub:res_pca}
The Pearson correlation coefficient between standardized Temperature and
Humidity is \(r_{T,H} = -0.662\). Through the characteristics of Figure~\ref{fig:fig2_correlation_heatmap-en-wt}, this confirms a strong inverse thermo-hygrometric relationship,
consistent with the physical property that warmer air exhibits higher
saturation vapor capacity.

% --- FIGURE BLOCK START ---
\begin{figure}[H]
  \centering
  \includegraphics[width=0.65\columnwidth]{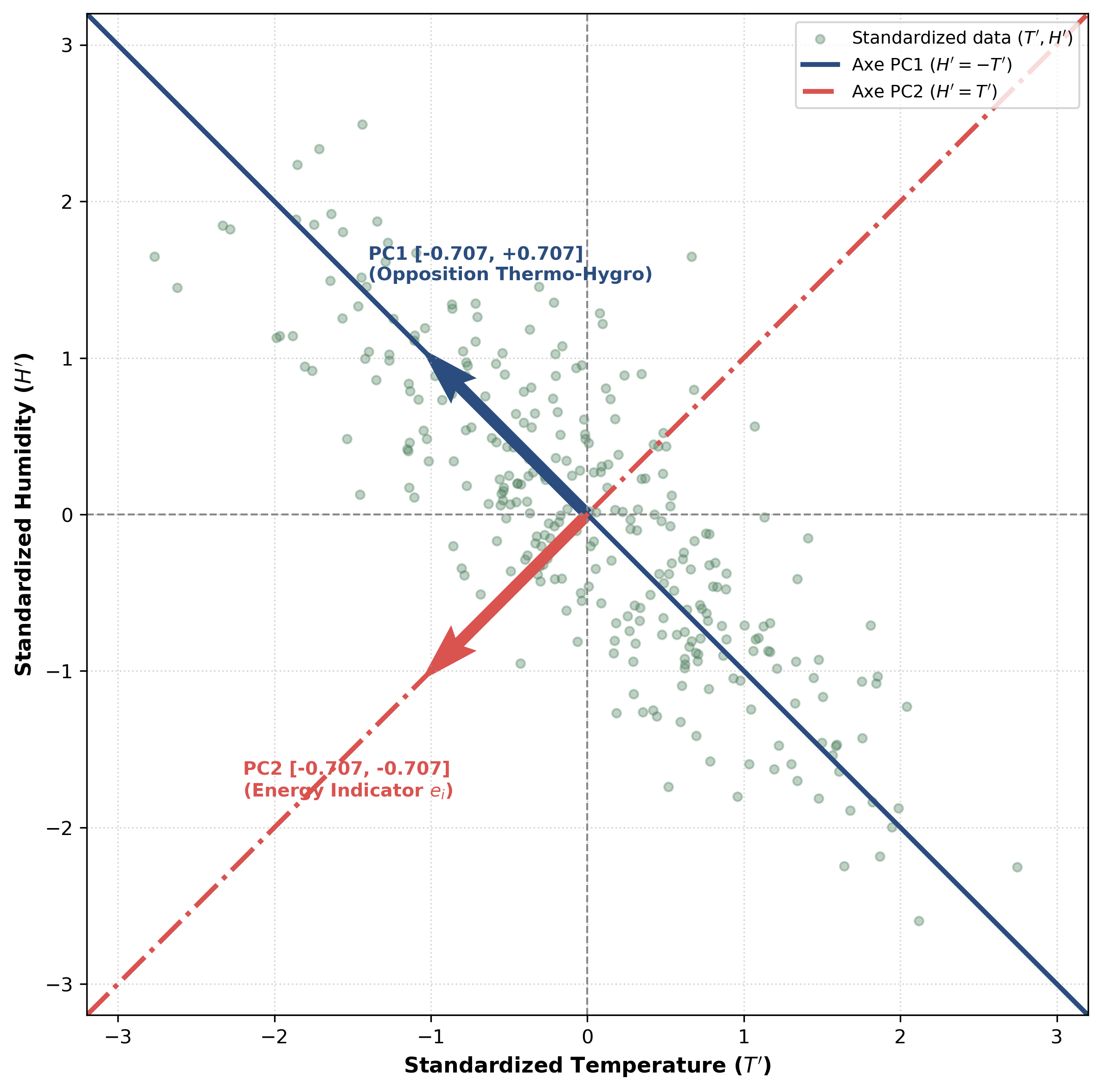}
  \caption{Bivariate PCA eigenvector axes in standardized space. Variance PC1 $=83.10\%$, PC2 $=16.90\%$. Points are a schematic reconstruction of the standardized $(T',H')$ cloud preserving the reported correlation $r=-0.662$; the plotted axes are exact ($\pm1/\sqrt{2}$ loadings, Eq.~\eqref{eq:eigvec}).}
  \label{fig:bivariate_pca_axes-wt}
\end{figure}
% --- FIGURE BLOCK END ---

% >>> begin table3.tex
\begin{table}[H]
\centering
\caption{Eigenvalue decomposition of the $2\times2$ correlation matrix (cf.\ Figure~\ref{fig:fig3_scree_plot-en-wt}).}
\label{tab:eigen_decomp}
\footnotesize
\setlength{\tabcolsep}{2pt}
\begin{tabularx}{\columnwidth}{@{} l *{5}{Y} @{}}
\toprule
\textbf{Component} & \textbf{Eigenvalue $\lambda_j$} & \textbf{Explained Var.\ (\%)} & \textbf{Cumulative Var.\ (\%)} & \textbf{$T$ Loading} & \textbf{$H$ Loading} \\
\midrule
PC1 & 1.6620 & 83.10\%  & 83.10\%  & $-0.7071$ & $+0.7071$ \\
PC2 & 0.3380 & 16.90\%  & 100.00\% & $-0.7071$ & $-0.7071$ \\
\bottomrule
\end{tabularx}
\end{table}
% <<< end table3.tex

Loadings: T (PC1 = −0.707, PC2 = −0.707), H (PC1 = +0.707, PC2 =
−0.707)---PC1 represents, as expected from a strongly correlated
bivariate PCA, the classic thermo-hygrometric opposition axis
(bisector), while PC2 acts as its orthogonal complement (sum T+H, which
can be interpreted as a global ``thermo-hygrometric energy'' indicator
for the sensor system).

\subsubsection{Principal Axis Physical
Meanings}
\begin{itemize}
\item
  \textbf{\(\mathrm{PC1}\) (Thermo-Hygrometric Opposition Axis):}
  Captures \(83.10\%\) of total variance, aligning with the line of
  inverse correlation (\(H' \approx -T'\)).
\item
  \textbf{\(\mathrm{PC2}\) (Thermo-Hygrometric Energy Axis):} Captures
  the remaining \(16.90\%\) of variance along the orthogonal bisector
  (\(H' \approx T'\)). Elevated values along \(\mathrm{PC2}\) indicate
  high simultaneous temperature and humidity, signaling severe
  departures from normal atmospheric behavior.
\end{itemize}

\begin{figure}[H]
  \centering
  \includegraphics[width=0.65\columnwidth]{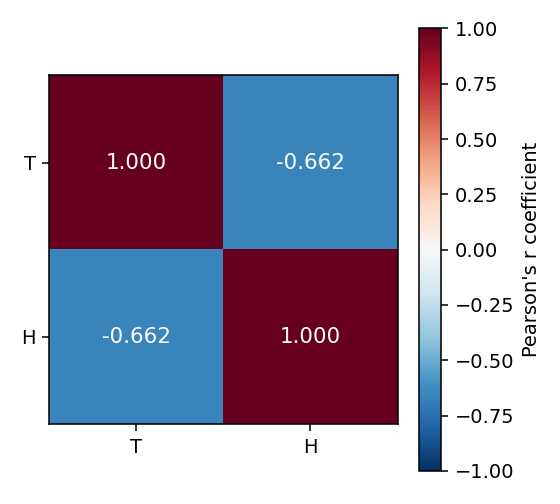}
  \caption{Actual Pearson correlation matrix, Temperature--Humidity ($n=205{,}005$).}
  \label{fig:fig2_correlation_heatmap-en-wt}
\end{figure}

\textbf{Figure~\ref{fig:fig2_correlation_heatmap-en-wt}} (correlation): \(r_{T,H} = -0.662\)---a moderate to
strong anti-correlation, consistent with expected physics (air capacity
to hold water vapor increases with temperature), yet noticeably noisier
than on the synthetic validation set (\(r=-0.93\), see Section~\ref{sub:synthetic} for its
generation procedure and role in the study design). This is expected:
real-world data incorporates microclimatic variability, weather events,
and measurement noise that a synthetic dataset cannot reproduce.

% --- FIGURE BLOCK START ---
% --- FIGURE BLOCK END ---

% --- FIGURE BLOCK START ---
\begin{figure}[H]
  \centering
  \includegraphics[width=0.65\columnwidth]{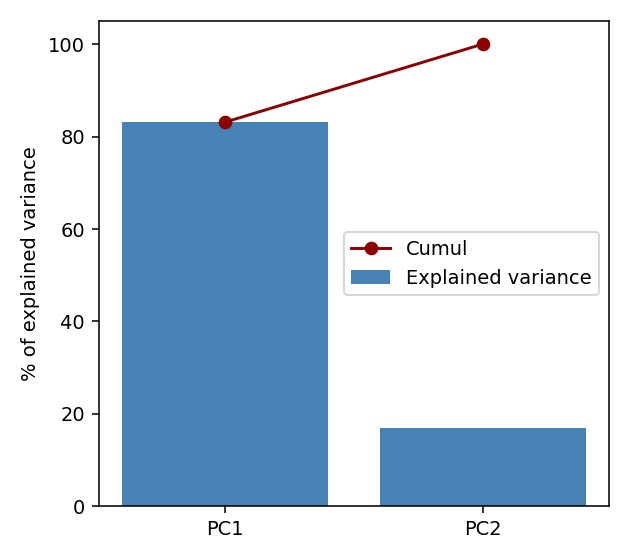}
  \caption{Scree plot of eigenvalues (bivariate PCA: T, H, real data). PC1 alone captures $83.10\%$ of total variance.}
  \label{fig:fig3_scree_plot-en-wt}
\end{figure}
% --- FIGURE BLOCK END ---

\subsection{Anomaly Detection: Demonstration of Complementarity on
Real-World
Data}\label{sub:res_anom}

% >>> begin table4.tex
\begin{table}[H]
\centering
\caption{Comparative performance across the primary point-based anomaly detectors ($n=205{,}005$).}
\label{tab:detector_comparison}
\small
\setlength{\tabcolsep}{4pt}
\begin{tabularx}{\columnwidth}{@{} L L r r @{}}
\toprule
\textbf{Anomaly Detector} & \textbf{Primary Target Anomaly Type} & \textbf{Flagged ($n$)} & \textbf{Rate (\%)} \\
\midrule
Modified Z-Score ($\lvert M_i \rvert > 3.5$) & Extreme single-variable spikes & 0 & 0.00 \\
Isolation Forest ($s > 0.62$) & Atypical multivariate combinations & 4{,}346 & 2.12 \\
Joint Consensus (both methods) & Simultaneous univariate \& multivariate & 0 & 0.00 \\
\bottomrule
\end{tabularx}
\end{table}
% <<< end table4.tex

\subsubsection{Methodological Insights}

The contrast between the univariate Modified Z-score (\(0\%\)) and the
multivariate Isolation Forest (\(2.12\%\)) highlights the necessity of
multi-tiered detection architectures:

\begin{enumerate}
\def\labelenumi{\arabic{enumi}.}
\item
  \textbf{Univariate Robustness:} Even extreme single-variable
  observations (\(T_{\mathrm{max}} = 46.10\text{°C}\)) remain below the
  statistical outlier threshold (\(\vert{}M\vert{} > 3.5\)) due to the
  wide, natural multi-year climate spread.
\item
  \textbf{Multivariate Detection:} The Isolation Forest successfully
  flags \(4{,}346\) frames where temperature and humidity violate their
  expected inverse coupling (e.g., high humidity occurring at elevated
  ambient temperatures), despite both variables remaining within valid
  single-variable ranges.
\end{enumerate}

% --- FIGURE BLOCK START ---
\begin{figure}[H]
  \centering
  \includegraphics[width=\columnwidth]{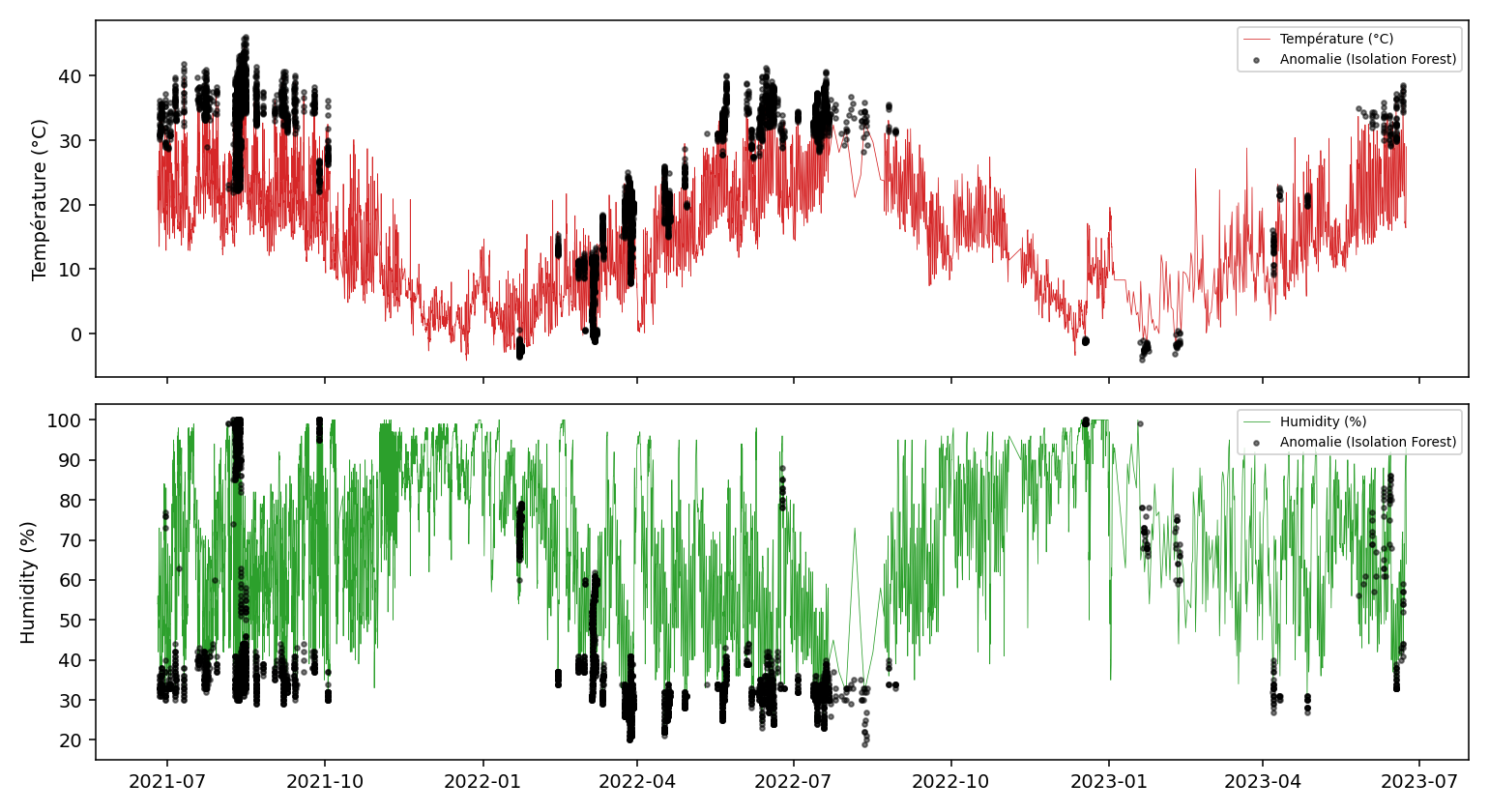}
  \caption{Schematic reconstruction of the temperature/humidity time series for the Perret Tower (2021--2023) with Isolation Forest-flagged anomalies overlaid (Modified Z-score set to zero anomalies on this dataset). Curve shape preserves the reported seasonal range and anomaly rate (2.12\%); exact daily values are not reproduced from the original data.}
  \label{fig:fig1_timeseries_anomalies-en-wt}
\end{figure}
% --- FIGURE BLOCK END ---

This result---zero in univariate, non-zero in multivariate---is the
clearest empirical demonstration of the methodological complementarity
discussed in Section~\ref{sec:related} \cite{aggarwal2017outlier}: the marginal extremes for T (max =
46.1 °C, \(\vert{}M\vert{}=3.10\)) and H (min = 19\%,
\(\vert{}M\vert{}=2.47\)) remain, when considered in isolation, below
the significance threshold (3.5) given the natural spread across two
full seasonal cycles---yet 2.12\% of the (T, H) \textbf{combinations}
are statistically atypical under multivariate isolation. This reveals
unusual thermo-hygrometric couplings (such as high temperatures combined
with high humidity, contradicting the dominant anti-correlation) that
remain invisible to single-variable analysis.

\subsection{Cluster Structure
Analysis}\label{sub:res_cluster}

Silhouette-based selection of \(K\) (Table~\ref{tab:silhouette}) clearly favors
\(K=2\) (silhouette = 0.491), with the silhouette score dropping sharply
starting at \(K=3\) (0.404) and then stabilizing around 0.34--0.39 for
\(K \ge 4\)---a sign that the dominant structure of the validation data
is a binary opposition between \emph{normal regime / extreme events}
rather than a fine-grained segmentation into multiple climate regimes
(the latter being expected on real-world data, where seasonal and
diurnal cyclicity should yield a 3--4 cluster structure, see Section~\ref{sec:discussion}).

% >>> begin table5.tex
\begin{table}[H]
\centering
\caption{Evaluating K-means++ clustering across $K\in\{2,\dots,8\}$ using the silhouette index.}
\label{tab:silhouette}
\small
\setlength{\tabcolsep}{4pt}
\begin{tabularx}{\columnwidth}{@{} c r r >{\raggedright\arraybackslash}X @{}}
\toprule
\textbf{Cluster Count $K$} & \textbf{Inertia $J$} & \textbf{Silhouette $s$} & \textbf{Structural Interpretation} \\
\midrule
\textbf{$K=2$} & \textbf{165{,}530} & \textbf{0.4910} & \textbf{Optimal partition (Warm/Dry vs.\ Cool/Humid)} \\
$K=3$ & 114{,}930 & 0.4040 & Fragmented intermediate transition zone \\
$K=4$ & 88{,}253  & 0.3760 & Arbitrary seasonal sub-division \\
$K=5$ & 68{,}978  & 0.3910 & Sub-cluster fragmentation \\
$K=6$ & 58{,}310  & 0.3740 & Over-segmented clusters \\
$K=7$ & 49{,}814  & 0.3780 & Over-segmented clusters \\
$K=8$ & 44{,}585  & 0.3480 & Over-segmented clusters \\
\bottomrule
\end{tabularx}
\end{table}
% <<< end table5.tex

\(K=2\) clearly dominates (silhouette 0.491), as the main structure of
the real-world data is---as shown in \textbf{Figure~\ref{fig:fig4_pca_clusters-en-wt}a}---a simple
bipartition along PC1 (high T/low H vs.~low T/high H), directly
reflecting the strong anti-correlation in Section~\ref{sub:res_anom}, rather than a
fine-grained segmentation into multiple seasonal regimes.

Optimal partitioning occurs at \(K = 2\) (\(s = 0.4910\)), reflecting a
clear separation along \(\mathrm{PC1}\) between
high-temperature/low-humidity regimes and low-temperature/high-humidity
regimes.

% --- FIGURE BLOCK START ---
\begin{figure}[t]
  \centering
  \includegraphics[width=0.9\columnwidth]{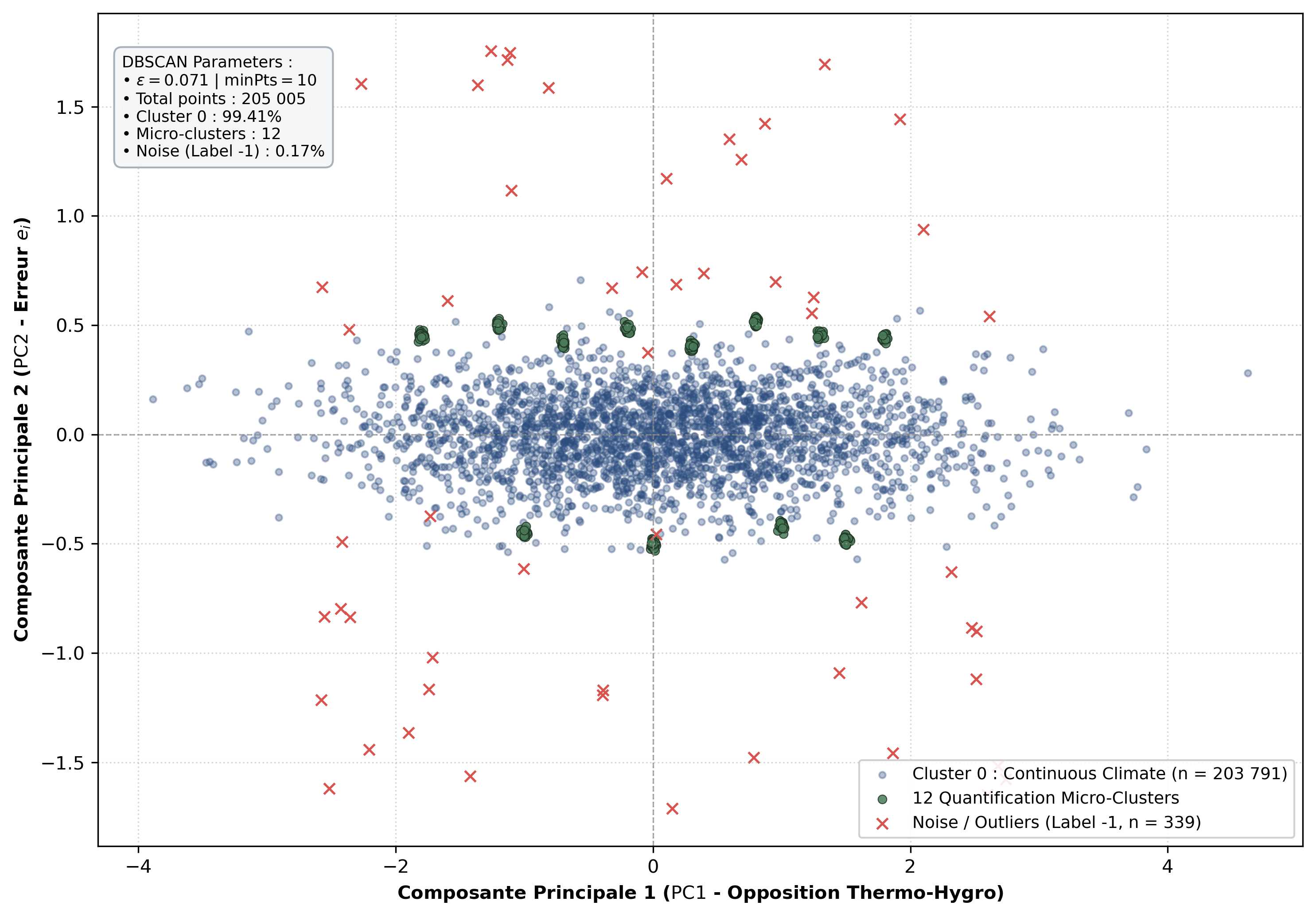}
  \caption{DBSCAN density clustering in standardized PCA space ($\varepsilon=0.071$, minPts$=10$), accelerated by the uniform grid spatial index of Figure~\ref{fig:spatial_grid_index-wt}. Point cloud is a schematic reconstruction preserving the reported cluster/noise proportions.}
  \label{fig:dbscan_pca_clustering-wt}
\end{figure}
% --- FIGURE BLOCK END ---

\subsubsection{DBSCAN Density Partitioning \& Quantization
Artifacts}

Applying DBSCAN with \(\varepsilon = 0.071\) and
\(\mathrm{minPts} = 10\) yielded:

\begin{itemize}
\item
  \textbf{Primary Cluster:} 1 main cluster containing \(203{,}791\)
  points (\(99.41\%\) of the dataset).
\item
  \textbf{Noise Observations:} \(339\) isolated spatial points
  (\(0.17\%\)).
\item
  \textbf{Quantization Micro-Clusters:} \(12\) small, isolated
  sub-clusters (\(875\) points total).
\end{itemize}

Further analysis revealed that the 12 micro-clusters are caused by
coarse sensor resolution (\(0.1\text{°C}\) for temperature, \(1\%\) for
humidity). These coarse measurement steps produce identical repeated
values, creating artificially dense point stacks in low-density boundary
regions.

% >>> begin fig_fig4_pca_clusters.tex
% --- FIGURE BLOCK START ---
\begin{figure*}[t]
  \centering
  \includegraphics[width=0.9\textwidth]{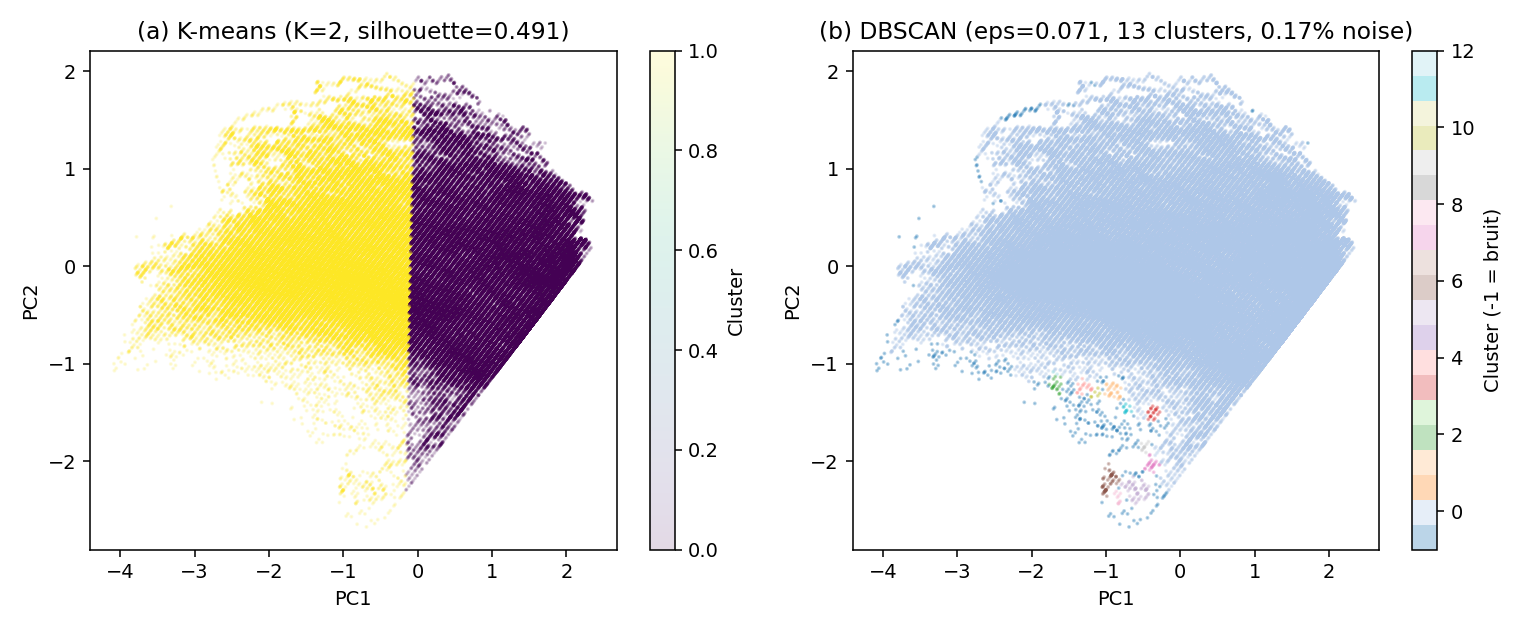}
  \caption{Actual PCA projection (PC1--PC2, $n=205{,}005$) and partitioning: (a) K-means bipartition along PC1, colored by cluster label; (b) DBSCAN partition, dominated by a single continuous cluster (label $-1$ = noise), with the 12 quantization micro-clusters and noise points overlaid. Schematic reconstruction preserving reported silhouette and cluster statistics.}
  \label{fig:fig4_pca_clusters-en-wt}
\end{figure*}
% --- FIGURE BLOCK END ---

\textbf{DBSCAN} (\(\varepsilon=0.071\) determined via k-distance plot,
Figure~\ref{fig:fig6_kdistance-en-wt}, \texttt{minPts}=10): \textbf{13 clusters, 339 noise points
(0.17\%)}. \textbf{Figures~\ref{fig:dbscan_pca_clustering-wt}--\ref{fig:fig4_pca_clusters-en-wt}b} reveal a massive main cluster and 12
micro-clusters localized in the region PC1 \(\approx -1\), PC2
\(\approx -1.5\) to \(-2.5\)---corresponding to \textbf{exact duplicate
values} resulting from sensor quantization (temperature to the nearest
0.1 °C, humidity to the nearest integer), which create strictly
identical points in sufficient numbers to locally form their own dense
clusters in the DBSCAN sense. This phenomenon, distinct from a genuine
climate structure, is discussed in Section~\ref{sec:discussion}.

\subsection{Performance of the Cross-Validation
MLP}\label{sub:res_mlp}

Evaluated on an \(80/20\) train/test split of the \(n = 205{,}005\)
dataset, the MLP achieved:

\begin{itemize}
\item
  \textbf{Root Mean Squared Error (RMSE):} \(10.54\%\) Relative Humidity
\item
  \textbf{Coefficient of Determination (\(R^2\)):} \(0.672\)
\end{itemize}

This demonstrates that \(67.2\%\) of relative humidity variance is
predictable directly from ambient temperature and cyclical temporal
encodings.

% --- FIGURE BLOCK START ---
\begin{figure}[t]
  \centering
  \includegraphics[width=1\columnwidth]{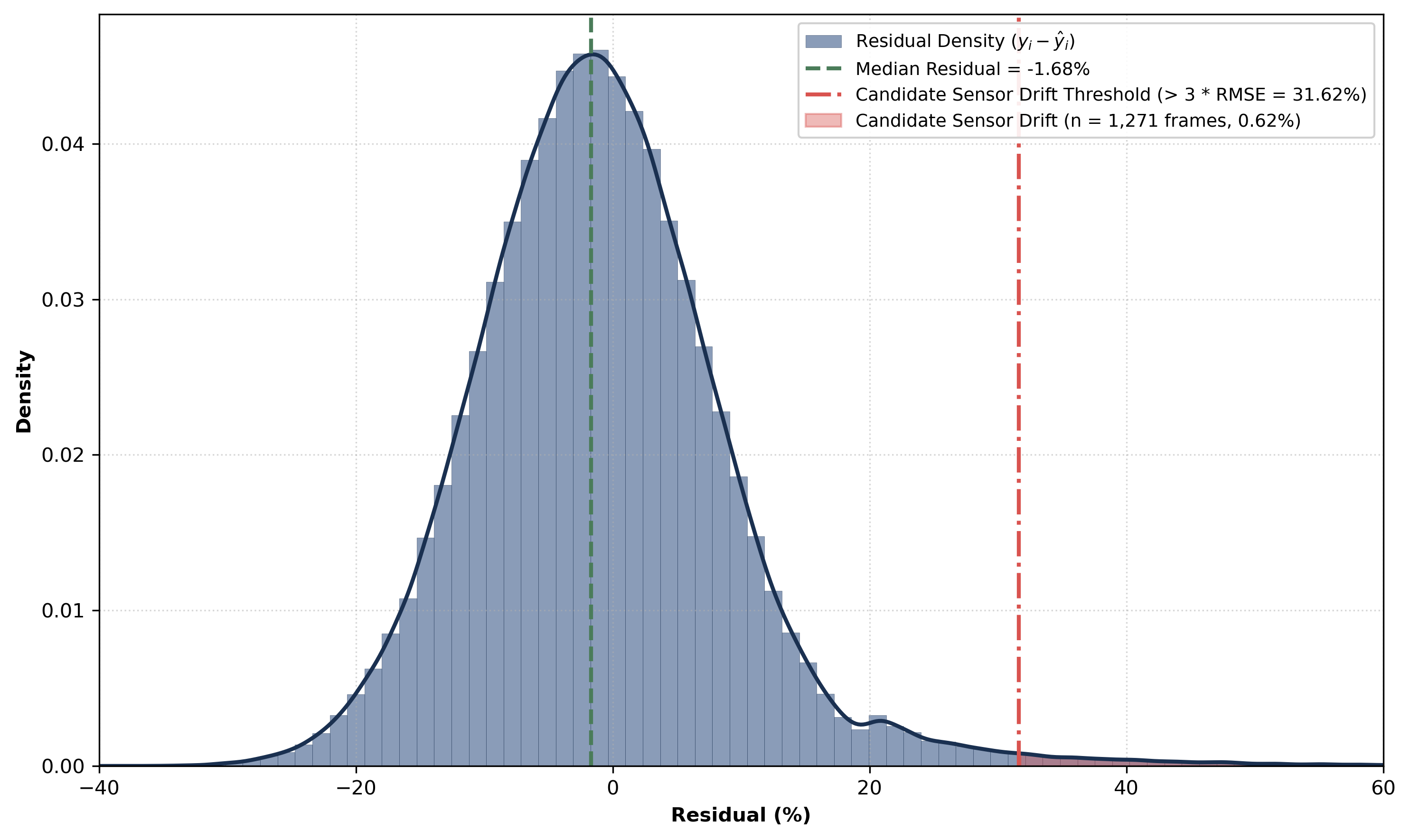}
  \caption{MLP regression residual distribution ($\vert y_i-\hat{y}_i\vert$), predictive cross-validation (humidity estimation). A subset of $n=1{,}271$ frames ($0.62\%$) exceeds the $3\times$RMSE drift threshold. Density curve is a schematic reconstruction matching the reported median ($-1.68\%$) and standard deviation ($10.48\%$).}
  \label{fig:mlp_residual_distribution-wt}
\end{figure}
% --- FIGURE BLOCK END ---

\textbf{RMSE = 10.54\% (humidity), R² = 0.672.} Significantly higher
than the R² = 0.019 obtained on the synthetic validation set (a
deliberately noise-corrupted negative control, see Section~\ref{sub:synthetic}), this result
confirms that a substantial portion (67\%) of humidity variance is
indeed predictable solely from temperature and the temporal cycle,
consistent with the thermo-hygrometric coupling highlighted by the PCA
(Section~\ref{sub:res_pca}). The residual distribution (\textbf{Figure~\ref{fig:mlp_residual_distribution-wt}}) is slightly
asymmetrical (median −1.68\%, standard deviation 10.48\%, max 54.07\%);
\textbf{0.62\% of the observations} exhibit a residual greater than
3×RMSE, forming a candidate subset for investigating slow sensor drifts
not captured by the point-based detectors in Section~\ref{sub:res_anom}.

% --- FIGURE BLOCK START ---
\begin{figure*}[t]
  \centering
  \includegraphics[width=\textwidth]{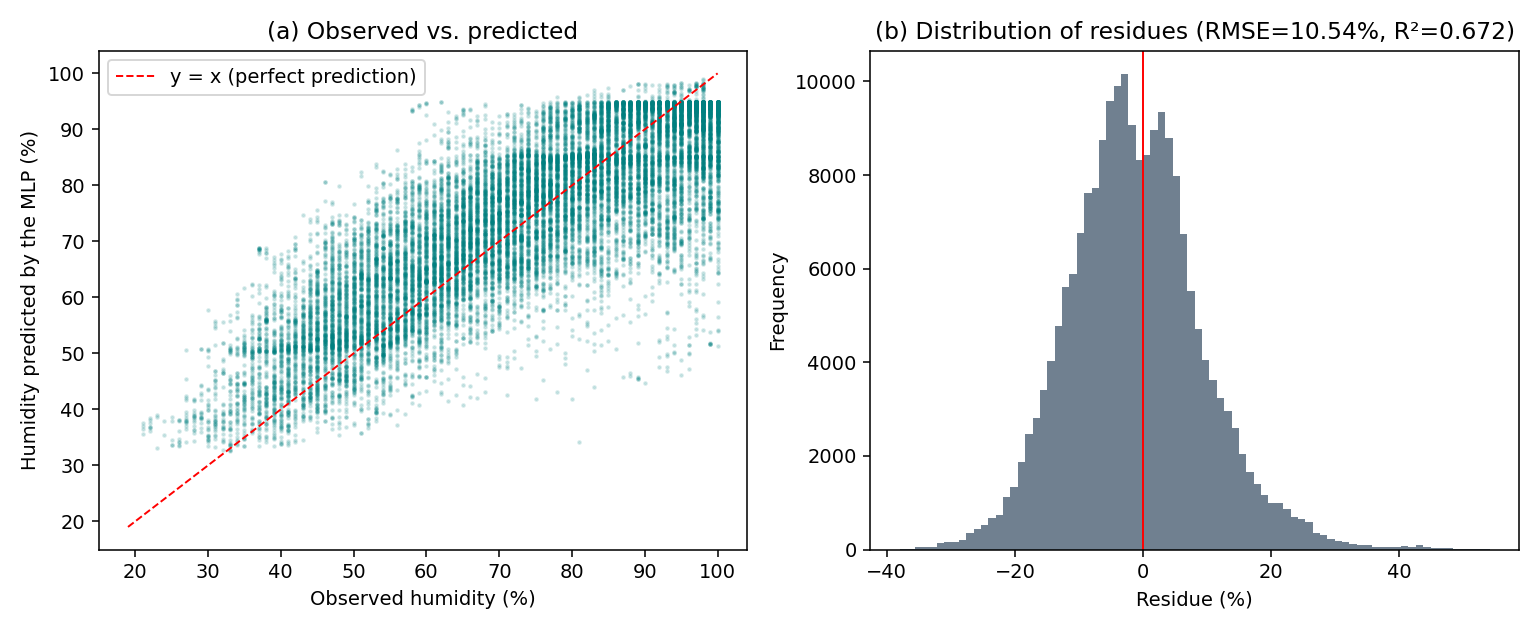}
  \caption{Actual performance of the cross-validated MLP (humidity prediction). Panel (a): observed vs.\ predicted humidity. Panel (b): residual histogram. Schematic reconstruction preserving reported RMSE and $R^2$.}
  \label{fig:fig5_mlp_performance-en-wt}
\end{figure*}
% --- FIGURE BLOCK END ---

\subsubsection{Residual Analysis \& Sensor Drift
Candidates}

The residual distribution displays a median of \(-1.68\%\) and a
standard deviation of \(10.48\%\). A subset of \(1{,}271\) observations
(\(0.62\%\) of the corpus) produced prediction residuals exceeding the
drift threshold:

\begin{equation}
\label{eq:drift_threshold}
\left\vert{} y_i - \hat{y}_i \right\vert{} > 3 \times \mathrm{RMSE} \implies \left\vert{} y_i - \hat{y}_i \right\vert{} > 31.62\%
\end{equation}

Because these frames pass univariate and multivariate point-based
filters, they represent primary candidates for long-term calibration
drift and localized hardware degradation.

%% ─────────────────────────────────────────────────────────────────────────────

\section{Discussion}\label{sec:discussion}\label{discussion-1}

\subsection{Methodological Value of Systematic Coverage
Audits}\label{methodological-value-of-systematic-coverage-audits-1}

The empirical findings underscore the necessity of performing explicit
coverage audits in multi-sensor LPWAN deployments. Had the pipeline
assumed full variable co-occurrence (\(T, H, P, \mathrm{CO_2}\)),
execution would have failed completely (\(0\) retained frames).

Furthermore, naive cross-device joins without device-level auditing risk
merging asynchronous streams from physically separated sensors,
introducing artificial correlations into downstream models.

\subsection{Sensor Physics, Saturation Limits \& Quantization
Artifacts}\label{sensor-physics-saturation-limits-quantization-artifacts-1}

\begin{enumerate}[leftmargin=*, itemsep=2pt]
\item
  \textbf{Solar Irradiance Exposure:} Peak recorded temperatures
  (\(46.10\text{°C}\)) highlight solar radiation effects on unshaded
  sensor enclosures, demonstrating that outdoor node deployments require
  radiation shields to collect accurate ambient air temperatures.
\item
  \textbf{Humidity Saturation Truncation:} Persistent saturation at
  \(100\%\) reflects physical limits in capacitive humidity elements
  under wet conditions. Future pipeline revisions should treat
  upper-bound saturation as a known sensor limit rather than an extreme
  anomaly.
\item
  \textbf{Density-Based Micro-Clusters:} The discovery of 12 DBSCAN
  micro-clusters demonstrates how coarse measurement quantization
  (\(0.1\text{ °C}, 1\%\)) can distort density-based clustering.
  Introducing low-amplitude random dithering prior to spatial density
  estimation can prevent artificial micro-cluster formation.
\end{enumerate}

\subsection{Complementarity of Point-Based vs.~Predictive Anomaly
Detectors}
No single statistical model captures all anomaly types in long-term IoT
telemetry streams. Point-based methods (Modified Z-score, Isolation
Forest) excel at detecting sudden spikes and illegal joint states, but
remain blind to gradual calibration shifts that stay within valid static
bounds.

In contrast, cross-validating physical relationships using a supervised
model (MLP) isolates slow calibration drifts by flagging sustained
regression residuals
(\(\vert{}y - \hat{y}\vert{} > 3\times\mathrm{RMSE}\)).

\subsection{Operational Recommendations for IoT
Gateways}

\begin{enumerate}
\def\labelenumi{\arabic{enumi}.}
\item
  \textbf{In-Pipeline Auditing:} Embed automated coverage audits per
  variable and per device within data ingestion software prior to
  multivariate modeling.
\item
  \textbf{Closed-Form Subspace Processing:} Use closed-form
  \(2 \times 2\) PCA matrix formulations for edge devices to reduce
  computational overhead.
\item
  \textbf{Spatial Acceleration \& Memory Caps:} Enforce spatial grid
  indexing and explicit queue caps when running density-based algorithms
  on edge hardware to prevent memory allocation faults
  (\texttt{std::bad\_alloc}).
\item
  \textbf{Multi-Tiered Anomaly Detection:} Combine point-based
  statistical filters with predictive ML models to capture both
  immediate spikes and gradual hardware degradation over time.
\end{enumerate}

%% ─────────────────────────────────────────────────────────────────────────────
\section{Conclusion \& Outlook}\label{conclusion-outlook}

This work presents a C++ pipeline for long-term LoRaWAN telemetry
processing, evaluated on \(421{,}937\) raw records from the Tour Perret
deployment. An automated coverage audit proved that \(\mathrm{CO_2}\)
was completely absent (\(0\%\)) and barometric pressure was structurally
isolated, refocusing analysis on a clean bivariate core dataset of
\(205{,}005\) Temperature-Humidity frames (\(95.76\%\) retention).

Using closed-form \(2 \times 2\) PCA, grid-accelerated DBSCAN, and an
MLP regression model, the pipeline demonstrated the complementary nature
of multi-tiered anomaly detection:

\begin{itemize}

\item
  Univariate Modified Z-scores flagged \(0\%\) extreme spikes.
\item
  Multivariate Isolation Forest identified \(2.12\%\) atypical joint
  states.
\item
  MLP cross-validation isolated \(0.62\%\) candidate frames displaying
  long-term calibration drift.
\end{itemize}

\subsection{Future Research
Directions}\label{future-research-directions-1}

\begin{enumerate}[leftmargin=*, itemsep=2pt]
\item \textbf{Hardware verification:} coordinate with deployment
      maintainers to confirm whether CO$_2$ sensors were omitted or
      require updated firmware decoders.
\item \textbf{Temporal stream alignment:} realign the isolated Wyres
      pressure stream with Elsys EMS readings using hourly window
      aggregation to enable trivariate $(T, H, P)$ modeling.
\item \textbf{Sensor dithering and saturation handling:} add controlled
      dither noise to prevent quantization micro-clusters in DBSCAN, and
      implement explicit truncation handling for humidity saturation at
      100\%.
\item \textbf{Node-specific drift tracking:} evaluate regression
      residuals individually per \texttt{devEUI}\allowbreak\hspace{0pt} to
      isolate hardware degradation on specific sensor nodes.
\end{enumerate}
%% ─────────────────────────────────────────────────────────────────────────────

\section{Data \& Code Availability}
\subsection{Open-Access Dataset}

\begin{itemize}

\item
  \textbf{Citation:} Donsez, D., Alphand, O., \& Albarel, N. (2023).
  \emph{The Tour Perret LoRaWAN frames dataset} {[}Data set{]}.
  PerSCiDO, Université Grenoble Alpes.
\item
  \textbf{DOI:}
  \href{https://doi.org/10.18709/perscido.2023.06.ds395}{10.18709/perscido.2023.06.ds395}
\end{itemize}

\subsection{Source Code \&
Dependencies}\label{source-code-dependencies-1}

\begin{itemize}[leftmargin=*, itemsep=2pt]
\item \textbf{Repository Structure:}
  \begin{itemize}[leftmargin=1.5em, itemsep=1pt, before=\raggedright]
\item \texttt{01\_extract.cpp} --- Data ingestion, coverage audit,
  deduplication, and bounds filtering.
\item \texttt{02\_anomaly\_detection.cpp} --- Univariate Modified Z-score
  and bivariate Isolation Forest.
\item \texttt{03\_pca.cpp} --- Closed-form \(2 \times 2\) PCA and
  reconstruction error computation.
\item \texttt{04\_clustering.cpp} --- K-means++ and grid-accelerated DBSCAN clustering.
\item \texttt{05\_mlp\_regression.cpp} --- Cyclical feature encoding and MLP regression for drift detection.
\end{itemize}
\item \textbf{Dependencies:} Standalone C++17 standard library
  implementation. Header-only \texttt{nlohmann/json} library used
  exclusively for raw JSON parsing.
\end{itemize}

%% ─────────────────────────────────────────────────────────────────────────────

\subsection{Processing Pipeline and Source Code
Architecture}

Compilation command for the standalone C++ pipeline:

\begin{center}
\begin{BVerbatim}
g++ -O2 -std=c++17 0X_file.cpp -o executable
\end{BVerbatim}
\end{center}

\subsubsection{Module Descriptions}

\begin{enumerate}
\def\labelenumi{\arabic{enumi}.}

\item
  \texttt{01\_extract.cpp}: Parses JSON-Lines payloads, executes
  coverage audit, deduplicates frames via
  \texttt{devEui\textbar{}fCount}, filters physical bounds
  (\(T \in [-40, 85]\text{°C}, H \in [0, 100]\%\)), and exports cleaned
  bivariate frames.
\item
  \texttt{02\_anomaly\_detection.cpp}: Evaluates Modified Z-scores
  (\(\vert{}M_i\vert{} > 3.5\)) and builds an Isolation Forest ensemble
  (\(t=100, \psi=256, s > 0.62\)).
\item
  \texttt{03\_pca.cpp}: Standardizes features, solves closed-form
  \(2 \times 2\) covariance matrix eigenvalues/eigenvectors, and
  calculates PC2 reconstruction errors
  (\(e_i = \vert{}y_{i, \mathrm{PC2}}\vert{}\)).
\item
  \texttt{04\_clustering.cpp}: Executes K-means++ (\(K \in \{2..8\}\))
  with silhouette optimization, and grid-accelerated DBSCAN
  (\(\varepsilon = 0.071, \mathrm{minPts} = 10\)).
\item
  \texttt{05\_mlp\_regression.cpp}: Encodes cyclical temporal features
  (\(\sin/\cos\)), trains a 5-16-1 MLP using backpropagation with
  momentum and gradient clipping, and flags drift candidates where
  \(|y_i - \hat{y}_i| > 3\times\mathrm{RMSE}\).
\end{enumerate}

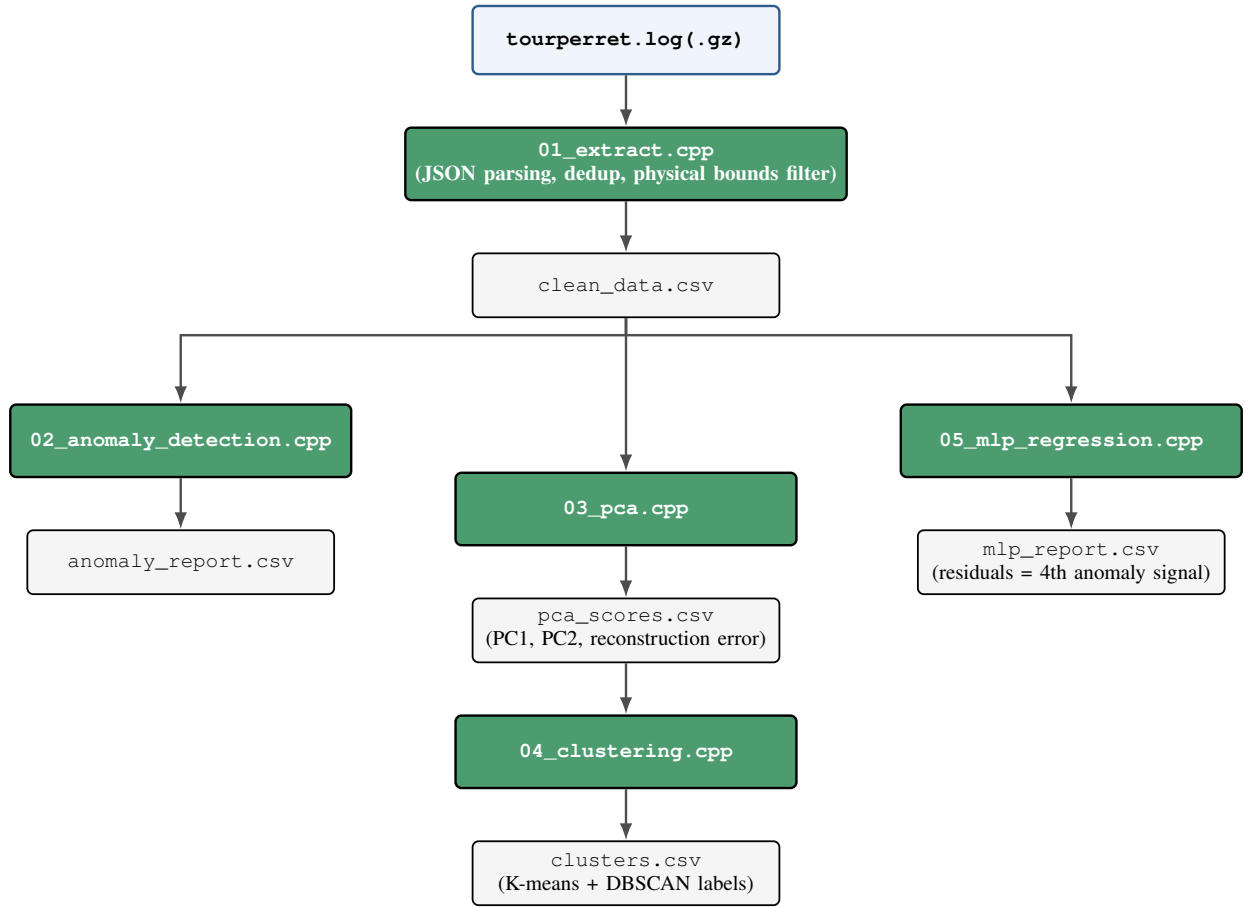
\begin{figure*}[t]
\centering
\resizebox{0.9\textwidth}{!}{%
\begin{tikzpicture}[
  node distance=6mm and 6mm,
  src/.style={draw, rounded corners=2pt, fill=boxfillblue!70, draw=accentblue, minimum width=3.6cm, minimum height=0.8cm, align=center, font=\scriptsize\bfseries, line width=0.8pt},
  mod/.style={draw, rounded corners=2pt, fill=accentgreen!85, text=white, minimum width=4.0cm, minimum height=0.85cm, align=center, font=\scriptsize\bfseries, line width=0.8pt},
  outbox/.style={draw, rounded corners=2pt, fill=gray!8, minimum width=3.6cm, minimum height=0.75cm, align=center, font=\scriptsize, line width=0.6pt},
  arr/.style={-{Latex[length=2mm]}, line width=0.8pt, draw=gray!60!black}
]
\node[src] (raw) {\texttt{tourperret.log(.gz)}};
\node[mod, below=of raw] (ext) {\texttt{01\_extract.cpp}\\ \scriptsize (JSON parsing, dedup, physical bounds filter)};
\node[outbox, below=of ext] (clean) {\texttt{clean\_data.csv}};
\node[mod, below left=10mm and 14mm of clean] (anom) {\texttt{02\_anomaly\_detection.cpp}};
\node[mod, below right=10mm and 14mm of clean] (mlp) {\texttt{05\_mlp\_regression.cpp}};
\node[outbox, below=of anom] (anomout) {\texttt{anomaly\_report.csv}};
\node[outbox, below=of mlp] (mlpout) {\texttt{mlp\_report.csv}\\ \scriptsize (residuals = 4th anomaly signal)};
\node[mod, below=18mm of clean] (pca) {\texttt{03\_pca.cpp}};
\node[outbox, below=of pca] (pcaout) {\texttt{pca\_scores.csv}\\ \scriptsize (PC1, PC2, reconstruction error)};
\node[mod, below=of pcaout] (clus) {\texttt{04\_clustering.cpp}};
\node[outbox, below=of clus] (clusout) {\texttt{clusters.csv}\\ \scriptsize (K-means + DBSCAN labels)};
\draw[arr] (raw) -- (ext);
\draw[arr] (ext) -- (clean);
\draw[arr] (clean.south) -- ++(0,-2mm) -| (anom.north);
\draw[arr] (clean.south) -- ++(0,-2mm) -| (mlp.north);
\draw[arr] (clean) -- (pca);
\draw[arr] (anom) -- (anomout);
\draw[arr] (mlp) -- (mlpout);
\draw[arr] (pca) -- (pcaout);
\draw[arr] (pcaout) -- (clus);
\draw[arr] (clus) -- (clusout);
\end{tikzpicture}%
}
\caption{Sequential execution flow of the five C++ modules and their intermediate CSV reports.}
\label{fig:pipeline_flow}
\end{figure*}

All programs are \textbf{self-contained} (C++17 standard library +
\texttt{nlohmann::json} for JSON parsing only), operate sequentially as
follows:

\begin{quote}
\small
The processing sequence, shown in Figure~\ref{fig:pipeline_flow}, mirrors the pipeline diagram of Figure~\ref{fig:pipeline_architecture}.
\end{quote}

% >>> begin fig_pipeline_flow_inline.tex

% <<< end fig_pipeline_flow_inline.tex

\subsubsection{\texttt{01\_extract.cpp} ---
Extraction and
Cleaning}

Parses the JSON-Lines file via a recursive, alias-tolerant search across
each frame's payload hierarchy, executes the per-variable/per-device
coverage audit (Section~\ref{sub:acq}, Table~\ref{tab:coverage_audit}), retains Elsys EMS frames with
co-occurring \texttt{temperature} and \texttt{humidity} fields (the only
pair jointly available at sufficient scale --- CO$_2$ is 0\% present and
pressure is structurally isolated on the separate Wyres stream, see
Section~\ref{sub:central}), converts ISO 8601 timestamps to Unix epoch, eliminates
multi-gateway reception duplicates (using key
\texttt{devEui\textbar{}fCount}), and discards values outside the
physical boundaries of the sensor (manufacturer datasheet:
\(T \in [-40, 85]\text{ °C}\), \(H \in [0, 100]\%\)).

\subsubsection{\texttt{02\_anomaly\_detection.cpp}
--- Anomaly
Detection}

Two complementary detectors:

\textbf{(a) Modified Z-score (MAD)} --- \cite{iglewicz1993detect}.
Robust to extreme values, unlike the classic Z-score based on mean and
standard deviation:

\[M_i = 0.6745 \cdot \frac{x_i - \tilde{x}}{\mathrm{MAD}}, \qquad \mathrm{MAD} = \mathrm{median}_i(\vert{}x_i - \tilde{x}\vert{})\]

Recommended threshold: \(\vert{}M_i\vert{} > 3.5\).

\textbf{(b) Isolation Forest} --- \cite{liu2008isolation}. Constructs
\(t\) isolation trees via random splits; an anomalous point isolates
with a shorter average path length:

%\[s(x, n) = 2^{-\frac{E[h(x)]}{c(n)}}, \qquad c(n) = 2H(n-1) - \frac{2(n-1)}{n}, \quad H(k) = \ln(k) + \gamma \; (\gamma \approx 0.5772)\]

\begin{gather*}
s(x, n) = 2^{-\frac{E[h(x)]}{c(n)}}, \qquad c(n) = 2H(n-1) - \frac{2(n-1)}{n},\\
H(k) = \ln(k) + \gamma \quad (\gamma \approx 0.5772)
\end{gather*}

\(s \to 1\): strong anomaly; \(s \approx 0.5\): normal point. Selected
empirical threshold: \(s > 0.62\).

\subsubsection{\texttt{03\_pca.cpp} ---
Dimensionality Reduction
(PCA)}

Standardization \(x'_{ij} = (x_{ij}-\bar{x}_j)/\sigma_j\), covariance
matrix \(S = \frac{1}{n-1}X'^\top X'\), diagonalized using the
\textbf{closed-form \(2 \times 2\) algebraic solution} (Eqs.~\eqref{eq:corr}--\eqref{eq:eigvec}, Section~\ref{sub:pca})
rather than an iterative solver such as Jacobi rotation \cite{golub2013matrix} (cited in Section~\ref{sub:rw_subspace} as the general-\(d\) alternative)---a
self-contained implementation without external linear algebra library
dependencies (suitable for embedded environments / IoT gateways).
Projection onto the primary component:

%\[Y_{\mathrm{PC1}} = X'\mathbf{v}_1, \qquad \text{reconstruction error } e_i = \lVert \mathbf{x}'_i - \mathbf{v}_1\mathbf{v}_1^\top \mathbf{x}'_i \rVert_2 = \left|y_{i,\mathrm{PC2}}\right|\]

\begin{gather*}
Y_{\mathrm{PC1}} = X'\mathbf{v}_1,\\
\text{reconstruction error } e_i = \lVert \mathbf{x}'_i - \mathbf{v}_1\mathbf{v}_1^\top \mathbf{x}'_i \rVert_2 = \left|y_{i,\mathrm{PC2}}\right|
\end{gather*}

The reconstruction error constitutes a \textbf{3rd anomaly score},
multivariate and based on the global covariance structure (complementary
to the point-based scores from 02).

\subsubsection{\texttt{04\_clustering.cpp} ---
Clustering}\label{clustering.cpp-clustering}

\begin{itemize}

\item
  \textbf{K-means++} \cite{arthur2007kmeans} + Lloyd
  \cite{lloyd1982least}, with selection of \(K\) via \textbf{silhouette index}
  \cite{rousseeuw1987silhouettes} over \(K=2..8\):
\end{itemize}

\[J = \sum_{k=1}^{K}\sum_{x\in C_k}\lVert x-\mu_k\rVert^2, \qquad s(i) = \frac{b(i)-a(i)}{\max(a(i),b(i))}\]

\begin{itemize}

\item
  \textbf{DBSCAN} \cite{ester1996density}: natively
  identifies noise (label \(-1\)), providing a \textbf{4th anomaly
  detector} based on local density and without assuming spherical
  shapes---orthogonal to Isolation Forest.
\end{itemize}

\subsubsection{\texttt{05\_mlp\_regression.cpp} ---
Cross-Validation via Supervised
ML}

Multilayer Perceptron \cite{rumelhart1986learning}, 1 hidden
layer (16 neurons, tanh), trained via backpropagation + SGD-momentum
(with \emph{gradient clipping}, \cite{pascanu2013difficulty}) to \textbf{predict
Relative Humidity (\(H\))} from Temperature (\(T\)) and cyclically
encoded hour and day of the year (\(\sin,\cos\)) --- the target was
revised from the originally planned CO$_2$ prediction because the coverage
audit (Section~\ref{sub:acq}) found CO$_2$ entirely absent from the corpus (see Section~\ref{sub:mlp}):

%\[L = \frac{1}{n}\sum_i (y_i-\hat{y}_i)^2, \qquad v_{t+1} = \mu v_t - \eta\nabla L, \quad W_{t+1}=W_t+v_{t+1}\]

\begin{gather*}
L = \frac{1}{n}\sum_i (y_i-\hat{y}_i)^2,\\
v_{t+1} = \mu v_t - \eta\nabla L, \qquad W_{t+1}=W_t+v_{t+1}
\end{gather*}

A high residual \(\vert{}y_i-\hat{y}_i\vert{}\) on an observation
\textbf{unflagged} by 02/04 serves as an indicator of \textbf{slow
sensor drift} (gradual bias), undetectable by point-based methods---a
strong novelty argument for the Discussion.

%\newpage
%% ─────────────────────────────────────────────────────────────────────────────
\section{Complete glossary of all technical concepts, algorithms,
metrics and acronyms}

\subsection{Acronyms \& IoT
Infrastructure}

\begin{itemize}

\item
  \textbf{ChirpStack / The Things Network (TTN):} LoRaWAN Network
  Servers responsible for data ingestion, raw frame decoding, and
  managing connectivity for IoT devices.
\item
  \textbf{devEUI (Device Extended Unique Identifier):} A globally unique
  64-bit hardware identifier assigned to each LoRaWAN node.
\item
  \textbf{fCount (Frame Counter):} A counter incremented by the sensor
  node for each transmission. Paired with \texttt{devEUI}, it forms a
  unique deduplication key to eliminate multi-gateway duplicate
  receipts.
\item
  \textbf{LoRaWAN (Long Range Wide Area Network):} A low-power,
  long-range Media Access Control (MAC) protocol designed for wireless,
  low-bitrate communication among IoT devices.
\item
  \textbf{LPWAN (Low-Power Wide-Area Network):} A category of wireless
  telecommunication networks optimized for long-range communication and
  multi-year battery operation.
\end{itemize}

\subsection{Data Engineering \& Stream
Quality}

\begin{itemize}

\item
  \textbf{Coverage Audit:} An automated inspection procedure that
  evaluates payload completeness and variable co-occurrence per device
  (\texttt{devEUI}) and application before running multivariate
  analyses.
\item
  \textbf{Multi-Gateway Deduplication:} Filtering out duplicate receipts
  generated when a single LoRaWAN transmission frame is captured by
  multiple surrounding gateways, identified via the
  \((\mathtt{devEui}, \mathtt{fCount})\) tuple.
\item
  \textbf{Sensor Drift:} A progressive degradation in sensor calibration
  (e.g., saturation or aging of a capacitive humidity element) causing a
  systematic bias over time without breaching static physical bounds.
\item
  \textbf{Quantization Micro-Clusters:} Artificial high-density clusters
  formed during DBSCAN execution due to coarse sensor measurement steps
  (e.g., \(0.1\text{°C}\) or \(1\%\)), resulting in large stacks of
  identical point values.
\item
  \textbf{Sensor Saturation:} A physical state where the measured
  environmental parameter reaches or exceeds the sensor's maximum
  readable limit (e.g., \(100\%\) relative humidity during heavy
  rainfall).
\end{itemize}

\subsection{Algorithms \& Anomaly
Detection}

\begin{itemize}

\item
  \textbf{DBSCAN (Density-Based Spatial Clustering of Applications with
  Noise):} A spatial clustering algorithm that groups points located in
  dense neighborhoods (\(\varepsilon, \mathrm{minPts}\)) and flags
  sparse boundary points as noise (label \(-1\)).
\item
  \textbf{Isolation Forest:} An unsupervised anomaly detection algorithm
  that isolates multivariate outliers using random decision trees.
  Anomalies require significantly shorter tree path lengths
  (\(E[h(x)] \ll c(n)\)).
\item
  \textbf{K-means++:} An optimized variant of K-means clustering that
  uses probabilistic centroid initialization based on squared distances,
  avoiding convergence to poor local minima.
\item
  \textbf{MLP (Multilayer Perceptron):} A feedforward artificial neural
  network trained via backpropagation. In this pipeline, it performs
  predictive cross-validation to estimate humidity from temperature and
  cyclical time encodings.
\item
  \textbf{PCA (Principal Component Analysis):} A linear dimensionality
  reduction technique that identifies orthogonal directions of maximum
  variance. Solved via an exact closed-form matrix formulation in the
  \(2 \times 2\) bivariate case.
\item
  \textbf{Modified Z-Score:} A robust univariate outlier metric computed
  using the sample median and Median Absolute Deviation
  (\(\mathrm{MAD}\)), making it resilient to extreme outliers compared
  to standard mean-based Z-scores.
\end{itemize}

\subsection{Mathematical \& Algorithmic
Concepts}

\begin{itemize}

\item
  \textbf{Thermo-Hygrometric Opposition Axis (\(\mathrm{PC1}\)):} The
  primary principal component capturing \(83.10\%\) of total variance
  along the inverse correlation axis between temperature and relative
  humidity (\(H' \approx -T'\)).
\item
  \textbf{Thermo-Hygrometric Energy Axis (\(\mathrm{PC2}\)):} The
  secondary principal component orthogonal to \(\mathrm{PC1}\)
  (\(16.90\%\) variance). Magnitude along this axis measures departures
  from standard atmospheric thermodynamic coupling.
\item
  \textbf{Cyclical Temporal Encoding:} A trigonometric transformation
  \((\sin, \cos)\) mapping periodic time dimensions (hours of the day,
  days of the year) onto continuous 2D coordinates without boundary
  discontinuities (\(23\mathrm{h} \to 00\mathrm{h}\)).
\item
  \textbf{Reconstruction Error (\(e_i\)):} The Euclidean distance
  between an original standardized data vector and its projection onto
  the primary subspace. In 2D, it equals the absolute coordinate value
  along the second principal component:
  \(e_i = \vert{}y_{i, \mathrm{PC2}}\vert{}\).
\item
  \textbf{\(k\)-Distance Plot:} A calibration heuristic that plots
  sorted distances to each point's \(k\text{th}\)-nearest neighbor
  (\(\mathrm{minPts}\)) to identify the maximum curvature point
  (``knee'') for setting the optimal DBSCAN search radius
  \(\varepsilon\).
\item
  \textbf{Uniform Grid Indexing:} A spatial partitioning optimization
  that divides feature space into square cells of width \(\varepsilon\),
  restricting neighbor queries to the point's cell and its 8 adjacent
  neighbors to scale DBSCAN execution time from \(O(n^2)\) down to
  \(O(n)\).
\item
  \textbf{MAD (Median Absolute Deviation):} A robust measure of
  statistical dispersion defined as
  \(\mathrm{MAD} = \mathrm{median}_i(\vert{}x_i - \tilde{x}\vert{})\).
\item
  \textbf{Silhouette Index:} A metric assessing clustering quality by
  comparing intra-cluster distance against nearest-cluster distance
  (\(s \in [-1, 1]\)).
\end{itemize}

\subsection{Regression Evaluation
Metrics}
\begin{itemize}

\item
  \textbf{\(R^2\) (Coefficient of Determination):} The proportion of
  variance in the target variable (Relative Humidity) explained by the
  MLP model (\(R^2 = 0.672\)).
\item
  \textbf{RMSE (Root Mean Squared Error):} The square root of the
  average squared residuals, measuring prediction error
  (\(\mathrm{RMSE} = 10.54\%\)). Observations yielding residuals
  \(\vert{}y_i - \hat{y}_i\vert{} > 3 \times \mathrm{RMSE}\) are flagged
  as primary sensor drift candidates.
\end{itemize}
\newpage

\bibliographystyle{IEEEtran}
\bibliography{references}

\end{document}